\documentclass{article} 
\usepackage{conference,times}

\usepackage{amsmath,amsfonts,bm}

\def\eqref#1{equation~\ref{#1}}

\def\1{\bm{1}}

\DeclareMathAlphabet{\mathsfit}{\encodingdefault}{\sfdefault}{m}{sl}
\SetMathAlphabet{\mathsfit}{bold}{\encodingdefault}{\sfdefault}{bx}{n}

\iclrfinalcopy

\usepackage{setspace}
\usepackage[utf8]{inputenc} 
\usepackage[T1]{fontenc}    
\usepackage[pagebackref,breaklinks,colorlinks]{hyperref}
\usepackage{url}            
\usepackage{booktabs}       
\usepackage{amsfonts}       
\usepackage{nicefrac}       
\usepackage{microtype}      
\usepackage{xcolor}         
\usepackage{graphicx} 
\usepackage{tabularx}
\usepackage{amsmath}
\usepackage{xspace}
\usepackage{graphbox}
\usepackage{float}
\usepackage[normalem]{ulem}
\usepackage{multirow}
\usepackage{caption}
\usepackage{wrapfig}

\newcommand{\OM}{GS-VTON}
\newcommand{\OMO}{GS-VTON }
\newcommand{\ie}{\textit{i.e.}\xspace}

\newcommand\pl[1]{{\color{black}#1}}
\newcommand\ToDo[1]{{\color{black}#1}}
\title{GS-VTON: Controllable 3D Virtual Try-on with Gaussian Splatting}

\author{Yukang Cao$^1$\thanks{Equal contribution \quad $^\dagger$ Corresponding authors} \quad Masoud Hadi$^{3*}$ \quad  Liang Pan$^{2\dagger}$ \quad  Ziwei Liu$^{1\dagger}$\\
$^1$S-Lab, Nanyang Technological University, $^2$Shanghai AI Laboratory, $^3$Isfahan University of Technology\\
\url{https://yukangcao.github.io/GS-VTON/}
}

\begin{document}

\maketitle

\begin{abstract}

\pl{
Diffusion-based 2D virtual try-on (VTON) techniques have recently demonstrated strong performance, while the development of 3D VTON has largely lagged behind.
Despite recent advances in text-guided 3D scene editing, integrating 2D VTON into these pipelines to achieve vivid 3D VTON remains challenging. 
The reasons are twofold.
First, text prompts cannot provide sufficient details in describing clothing.
Second, 2D VTON results generated from different viewpoints of the same 3D scene lack coherence and spatial relationships, hence frequently leading to appearance inconsistencies and geometric distortions.
To resolve these problems, we introduce an image-prompted 3D VTON method (dubbed \OM) which, by leveraging 3D Gaussian Splatting (3DGS) as the 3D representation, enables the transfer of pre-trained knowledge from 2D VTON models to 3D while improving cross-view consistency.
\textbf{(1)}~Specifically, we propose a personalized diffusion model that utilizes low-rank adaptation (LoRA) fine-tuning to incorporate personalized information into pre-trained 2D VTON models.
To achieve effective LoRA training, we introduce a reference-driven image editing approach that enables the simultaneous editing of multi-view images while ensuring consistency.
\textbf{(2)}~Furthermore, we propose a persona-aware 3DGS editing framework to facilitate effective editing while maintaining consistent cross-view appearance and high-quality 3D geometry.
\textbf{(3)}~Additionally, we have established a new 3D VTON benchmark, $\textit{3D-VTONBench}$, which facilitates comprehensive qualitative and quantitative 3D VTON evaluations.
Through extensive experiments and comparative analyses with existing methods, the proposed \OM has demonstrated superior fidelity and advanced editing capabilities, affirming its effectiveness for 3D VTON.
}

\end{abstract}
\section{Introduction}
\label{sec:intro}

Driven by advancements in neural rendering, virtual try-on (VTON) techniques represent a significant milestone in the intersection of fashion and computer vision.
These technologies are increasingly utilized across various domains, such as online shopping~\citep{kim2008adoption, zhang2019role}, VR/AR avatar modeling~\citep{mystakidis2022metaverse}, and gaming~\citep{lerner2007gaming}, enabling users to visualize how different garments will look on them without the need for a physical try-on.
Traditional methods~\citep{han2018viton, wang2018toward, meng2010interactive, hauswiesner2013virtual, hsieh2019fashionon} for this task primarily emphasize 2D image editing. Typically, they achieve virtual try-on by estimating pixel displacements using optical flow~\citep{canny1986computational} and employing pixel warping techniques to seamlessly blend clothing with the individual.
However, these 2D VTON approaches have struggled with occlusion issues and have difficulty accommodating complex human poses and clothing. 
With the rise of deep learning, methods~\citep{choi2021viton, ge2021disentangled, ge2021parser, lee2022high, men2020controllable} utilizing Generative Adversarial Networks (GANs)~\citep{goodfellow2014generative} have been introduced, aiming for more effective virtual fitting experiences. 
Despite their promise, these methods face challenges when handling custom user images that fall outside the training data. 
Although approaches~\citep{zhu2023tryondiffusion, choi2024improving, kim2024stableviton, xu2024ootdiffusion} leveraging large language models~\citep{radford2021learning} and diffusion models~\citep{song2021ddim, stable-diffusion} have demonstrated improved performance and generalization, these approaches still struggle with generating consistent multi-view images and accurately modeling 3D representations of garments.

Recently, neural radiance field (NeRF)~\citep{mildenhall2021nerf} and 3D Gaussian Splatting (3DGS)~\citep{kerbl20233d} have garnered significant attention for their efficient differentiable rendering capabilities, sparking research into text-guided 3D editing algorithms~\citep{haque2023instructnerf2nerf, ig2g, wu2024gaussctrl}.
Instruct-NeRF2NeRF~\citep{haque2023instructnerf2nerf} leverages a pre-trained diffusion model to edit rendered images while computing image-level loss based on textual prompts, allowing gradients to be back-propagated for modifying 3D differentiable scenes.
Following this, subsequent research efforts~\citep{zhuang2023dreameditor, shao2023control4d, dong2024vica, cheng2023progressive3d, han2023headsculpt, zhou2024headstudio} have aimed to improve quality and broaden the applications of Instruct-NeRF2NeRF across various tasks.
However, these methods generally apply global edits to the 3D scene, limiting their effectiveness for VTON applications.
While GaussianEditor~\citep{chen2023gaussianeditor} and TIP-Editor~\citep{zhuang2024tip} have been developed to facilitate local editing, they still encounter difficulties when modifying clothing items based solely on textual descriptions (see Fig.~\ref{fig:comparison}).
In addition, the rising use of image prompts in VTON applications, which convey richer information than text, underscores the urgent need for adaptable 3D VTON methods that accommodate user-specified images.
On the other hand,
directly applying 3D editing algorithms with diffusion-based 2D VTON models often leads to unsatisfactory results, primarily due to two major limitations.
\textit{First,} current 2D VTON diffusion models struggle to accurately visualize how the input clothing image would appear from different viewpoints, resulting in multi-view inconsistencies within the edited 3D scene. 
This issue stems from a lack of coherence and spatial relationships.
Furthermore, since we aim to modify individual garments rather than the entire body, maintaining consistency with other body parts becomes even more challenging. 
\textit{Second,} existing 2D VTON diffusion model may still yield suboptimal results when dealing with data that falls outside their training distribution, leading to issues such as blurriness and distortions in both appearance and geometry.

\begin{figure}[t]
  \centering
   \includegraphics[width=1\linewidth]{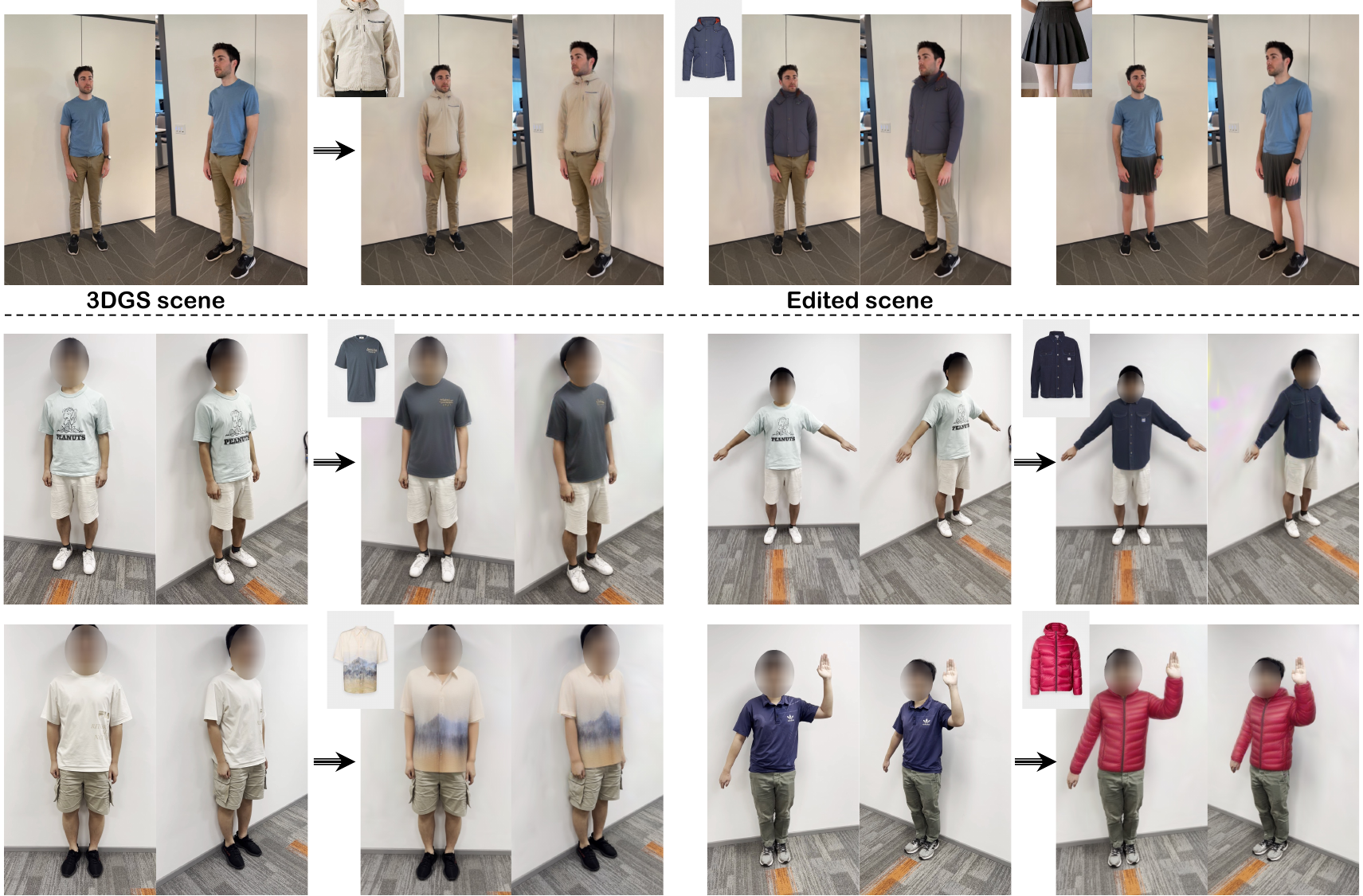}   
  \caption{\textbf{Examples of 3D virtual try-on results obtained via \OM.} Our approach facilitates high-fidelity editing of 3D garments, featuring intricate geometry and texture, under various scenarios with diverse cloth types, body shapes, and poses.}
   \label{fig:teaser}
\end{figure} 
To address this challenge, we present a novel image-prompted 3D VTON method in this paper, entitled \textbf{\OM}, which could achieve fine-grained editing of human garments. 
By taking a garment image and multi-view human images as input, our method comprises two major components, \ToDo{personalized diffusion model via LoRA fine-tuning and persona-aware 3DGS editing}, to achieve this objective.
\textit{First,} we enhance the pre-trained 2D VTON diffusion model by incorporating personalized information through a low-rank adaptation (LoRA) module. 
This enhancement allows the model to better reflect the specific characteristics of the input data by extending its learned distribution.
\textit{Second,} we introduce a reference-driven image editing approach that can simultaneously edit multi-view images 
while maintaining high consistency.
This method forms a robust foundation for effectively training the LoRA module.
\textit{Third,} we design a persona-aware 3DGS editing process that refines the original editing by blending two predicted attention features: one for editing and the other for ensuring coherence across different viewpoints. 
This strategy facilitates effective editing while enhancing multi-view consistency in geometry and texture.

Moreover, to support more thorough qualitative and quantitative evaluations, we establish a 3D VTON benchmark, named \textit{3D-VTONBench}, which, to our knowledge, is the first dataset of its kind.
As presented in Fig.~\ref{fig:teaser}, our method achieves high-fidelity 3D VTONs across diverse scenarios with various garments and human poses. 
Comprehensive comparisons with existing techniques also demonstrate that our approach significantly surpasses existing methods, establishing a new state-of-the-art in 3D VTON.

\ToDo{
Our contributions could be summarized as follows:
\begin{itemize}
    \item We introduce \OMO that, by extending the 2D pre-trained virtual try-on diffusion model to 3D, can take garment images as input to perform fine-grained 3D virtual try-on.
    \item To enhance multi-view consistency, we propose a reference-aware image editing technique that simultaneously generate consistent multi-view edited images, as well as a persona-aware 3DGS editing which takes into account both the intended editing direction and the original set of edited images.
    \item We have created the first benchmark for 3D virtual try-on, enabling more comprehensive evaluations. Extensive experiments demonstrate that our method establishes a new state-of-the-art performance for 3D virtual try-on.
\end{itemize}
}

\section{Related Works}

\paragraph{2D Diffusion-based Generative Model.}
In recent years, there have been significant advancements in vision-language technologies, including methods like Contrastive Language-Image Pretraining (CLIP)~\citep{radford2021clip} and various diffusion models~\citep{ho2020ddpm, dhariwal2021beatgan, rombach2022ldm,song2021ddim}. 
These models, trained on billions of text-image pairs, exhibit a strong understanding of real-world image distributions, enabling them to generate high-quality and diverse visuals. Such developments have greatly advanced the field of text-to-2D content generation~\citep{saharia2022photorealistic, ramesh2022dalle2, balaji2022ediffi, stable-diffusion, deepfloyd-if} and text-to-video generation~\citep{blattmann2023stable, liu2024sora, guo2023i2v, ma2024latte, huang2024vbench}.
Following these techniques, subsequent research has focused on enhancing control over generated outputs~\citep{zhang2023controlnet, zhao2023t2p, mou2023t2i-adapter}, adapting diffusion models for video sequences~\citep{singer2022make-a-video,blattmann2023videoldm}, facilitating both image and video editing~\citep{hertz2022prompt2prompt, kawar2022imagic, wu2022tuneavideo, brooks2022instructpix2pix,valevski2022unitune, esser2023gen-1,hertz2023dds}. Additionally, efforts have also been made to boost performance in personalized content generation~\citep{ruiz2022dreambooth, gal2022textual-inversion}.
Despite these advancements, the skill of crafting effective prompts remains crucial. Furthermore, in virtual try-on applications, which is the main target of this paper, textual descriptions frequently struggle to convey the intricate details of clothing as effectively as images, complicating the process of achieving realistic 2D virtual try-on.

\paragraph{Image-based Virtual Try-on.}
Image-based virtual try-on aims to create a visualization of a target person wearing a specific garment. Traditionally, methods~\citep{choi2021viton, lee2022high, men2020controllable, ge2021parser, xie2023gp, ge2021disentangled} based on generative adversarial network (GAN)~\citep{goodfellow2014generative} have been proposed to correspondingly deform the garment before fitting it to the human subject.  
Subsequent efforts~\citep{issenhuth2020not, lee2022high, ge2021parser, choi2021viton} have been made to minimize the discrepancies between the altered garment and the person.
However, these methods are often constrained by the training dataset, showing limited generalization to images outside the pre-trained distribution.
More recently, benefiting from the success of diffusion models~\citep{saharia2022photorealistic, ramesh2022dalle2, balaji2022ediffi, stable-diffusion}, researches have explored applying them to tackle the existing limitations for virtual try-on.
Specifically, TryOnDiffusion~\citep{zhu2023tryondiffusion} introduces a dual UNet architecture, demonstrating the potential of diffusion-based approaches when trained on extensive datasets;
\citet{yang2023paint} treats the virtual try-on as the exemplar-based image inpainting;
Stableviton~\citep{kim2024stableviton}, Ladi-VTON~\citep{morelli2023ladi} and \citet{gou2023taming} fine-tune diffusion models to achieve high-quality results;
IDM-VTON~\citep{choi2024improving} explores the usage of high-level semantics and low-level features to handle the task of identity preservation during virtual try-on.
Despite showing promise, they can still yield suboptimal results for out-of-distribution data, and transferring pre-trained 2D knowledge directly to the 3D space remains challenging.

\paragraph{3D Scene Editing.}
Leveraging the advancement of differentiable 3D representation, \ie, NeRF~\citep{mildenhall2020nerf} and 3DGS~\citep{kerbl20233d}, and diffusion-based text-to-2D generation methods~\citep{stable-diffusion, brooks2022instructpix2pix}, text-driven 3D scene editing methods have emerged for modifying 3D subjects using diffusion models.
Among them, Instruct-NeRF2NeRF (IN2N)~\citep{haque2023instructnerf2nerf} is the first to propose editing 2D renderings with Instruct-Pix2Pix~\citep{brooks2022instructpix2pix} and back-propagating gradients to adjust the 3D scene until convergence.
While IN2N shows promise, it faces challenges such as instability, inefficient training, blurry results, and significant artifacts. These issues arise from the diffusion models' lack of 3D awareness, particularly regarding camera pose, leading to inconsistent multi-view rendering edits.
To address these limitations, subsequent works~\citep{po2024state, wang2024survey} have aimed to enhance performance from various angles:
Instruct-Gaussian2Gaussian~\citep{ig2g} replaces the 3D representation of NeRF with 3DGS and introduces improved dataset updating strategies for better training efficiency.
Vica-NeRF~\citep{dong2024vica} first selects several reference images from the input dataset, edits them using Instruct-Pix2Pix, and then blends the results for the remaining dataset to reduce inconsistencies. However, this blending does not fully resolve the consistency issue and often results in blurry edits for human subjects.
DreamEditor~\citep{zhuang2023dreameditor} applies personalized DreamBooth~\citep{ruiz2023dreambooth} to achieve local editing.
TIP-Editor~\citep{zhuang2024tip} introduces a 3D bounding box as a condition to enhance control over local editing.
Despite promising results in adding objects to 3D scenes, these methods struggle with local modifications of internal geometry and textures.
GaussianEditor~\citep{chen2023gaussianeditor} utilizes large language models~\citep{kirillov2023segment} for text-driven local editing.
GaussCTRL achieves similar outcomes using a depth-conditioned ControlNet~\citep{zhang2023controlnet}.
Unfortunately, existing techniques typically do not accept images as input and have difficulty performing garment editing for effective 3D virtual try-on.
While GaussianVTON~\citep{chen2024gaussianvton} presents a three-stage editing pipeline aimed at a similar task, it may still face challenges in largely altering the original garment geometry.

\section{Methodology}
\label{sec:method}

\begin{figure}[t]
  \centering
   \includegraphics[width=1\linewidth]{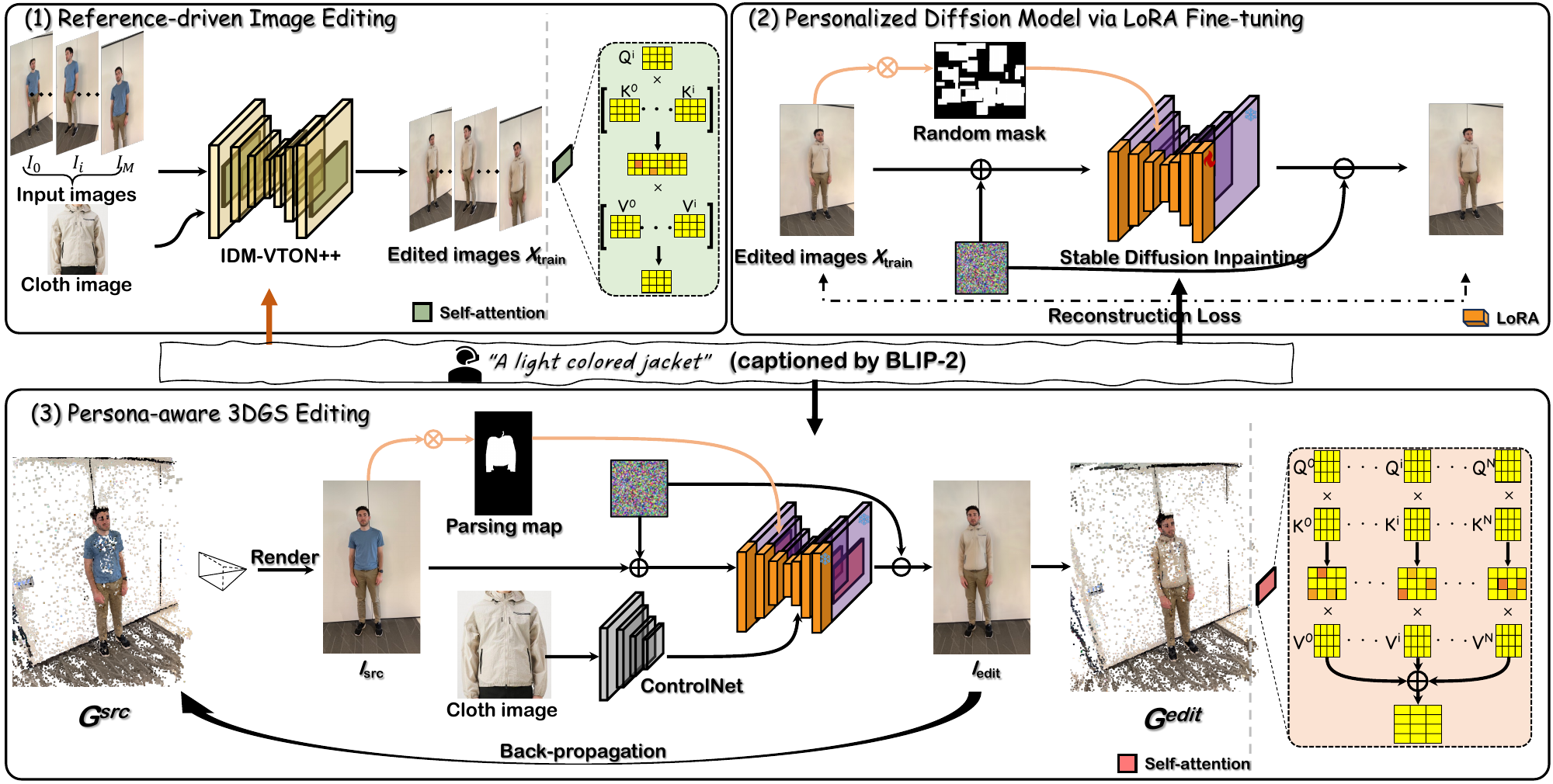}
  \caption{\textbf{Overview of \OM.} We enable 3D virtual try-on by leveraging knowledge from pre-trained 2D diffusion models and extending it into 3D space. \textbf{(1)} We introduce a reference-driven image editing method that facilitates consistent multi-view edits. \textbf{(2)} We utilize low-rank adaptation (LoRA) to develop a personalized inpainting diffusion model based on previously edited images. \textbf{(3)} The core of our network is the persona-aware 3DGS editing which, by leveraging the personalized diffusion model, respects two predicted attention features-one for editing and the other for ensuring coherence across different viewpoints-allowing for multi-view consistent 3D virtual try-on.}
   \label{fig:pipeline}
\end{figure}

We present \OM, a novel 3D virtual try-on method that enables controllable local editing to the human garment within a 3D Gaussian Splatting (3DGS) scene. 
Specifically, our method leverages multi-view human images $\mathcal{I}_{\text{train}}$, and a garment image as inputs to achieve this objective. 
In the subsequent sections, we first describe the preliminary knowledge that underpins our method in Sec.~\ref{sec:preliminaries}. We will then delve into the core elements of \OM, which include (1) personalized inpainting diffusion model adaptation via reference-driven image editing and LoRA fine-tuning in Sec.~\ref{sec:personalized_inpaiting}, and (2) persona-aware self-attention mechanism for achieving customizable 3D virtual try-ons using 3DGS in Sec.~\ref{sec:3dgs_editing}. An overview of \OMO is illustrated in Fig.~\ref{fig:pipeline}.

\subsection{Preliminaries}
\label{sec:preliminaries}

\paragraph{3D Gaussian Splatting.}
Unlike NeRF~\citep{mildenhall2021nerf}, which employs neural networks to synthesize novel views, 3D Gaussian Splatting (3DGS)~\citep{kerbl20233d} takes another direction by directly optimizing the 3D position $\mathbf{x}$ and attributes of 3D Gaussians, i.e, opacity $\alpha$, anisotropic covariance, and spherical harmonic (SH) coefficients $\mathcal{SH}$~\citep{ramamoorthi2001efficient}. Specifically, the 3D Gaussian $G(\mathbf{x})$ is defined by a 3D covariance matrix $\Sigma$ centered at point (mean) $\mu$:
\begin{equation}
    G(\mathbf{x})=e^{-\frac{1}{2}(\mathbf{x}-\mu)^T \Sigma^{-1}(\mathbf{x}-\mu)}.
\end{equation}

Drawing inspiration from~\citep{lassner2021pulsar}, 3DGS implements a tile-based rasterizer: The screen is first divided into tiles, such as $16 \times 16$ pixels. Each Gaussian is instantiated based on the number of tiles it overlaps, with a key assigned to each Gaussian to record view space depth and tile ID. These Gaussians are then sorted by depth, enabling the rasterizer to accurately manage occlusions and overlapping geometry. Finally, a point-based $\alpha$-blend rendering technique is used to compute the RGB color $\mathbf{C}$, by sampling points along the ray at intervals $\delta_i$:
\begin{equation}
    \mathbf{C}_{\text {color }}=\sum_{i \in N} \mathbf{c}_i \sigma_i \prod_{j=1}^{i-1}\left(1-\sigma_j\right), \quad \sigma_i = \alpha_i e^{-\frac{1}{2} (\mathbf{x})^\mathrm{T} \boldsymbol{\Sigma}^{-1} (\mathbf{x})},
\end{equation}
where $\mathbf{c}_i$ is the color of each point along the ray.

\paragraph{Instruct-Gaussian2Gaussian (IG2G)~\citep{ig2g}.}
Building on Instruct-Pix2Pix~\citep{brooks2022instructpix2pix} and 3DGS, IG2G facilitates text-guided scene editing with a given 3DGS model and its associated training dataset. This process is achieved in two main steps: 

\textit{1) Image editing.} 
For a rendered image from a specified camera viewpoint, IG2G first introduces Gaussian noise to the image. This noisy image, alongside the text embedding $y$ and the original training image, serves as conditions for Instruct-Pix2Pix to generate an edited image, which reflects the desired modifications. These changes will then be back-propagated to the 3DGS scene to update it accordingly.

\textit{2) Dataset update.} 
In addition to incorporating the editing direction through back-propagation, IG2G updates the entire dataset periodically, specifically every 2,500 training iterations. This update process involves inputting the rendered image into the diffusion model, such as Instruct-Pix2Pix, to ensure stronger and more accurate 3D edits over time.

\paragraph{Latent Diffusion Model.}
Latent Diffusion Model (LDM)~\citep{blattmann2023align} is a refined variant of diffusion models, optimizing the trade-off between image quality and training efficiency.
Specifically, LDM achieves this by first using a pre-trained variational auto-encoder (VAE)~\citep{kingma2013auto} to project images into a latent space, and then carry out the diffusion process in the latent space.
Additionally, LDM enhances the UNet architecture~\citep{ronneberger2015u} by incorporating self-attention mechanisms~\citep{vaswani2017attention}, cross-attention layers~\citep{vaswani2017attention}, and residual blocks~\citep{he2016deep}, allowing the model to integrate text prompts as conditional inputs during the image generation process. 
The attention mechanism in LDM's UNet is defined as follows:
\begin{equation}
\mathtt{ATT}(Q, K, V) = \text{softmax}(\frac{Q \cdot K^T}{\sqrt{d_k}}) \cdot V
\label{eq:att}
\end{equation}
where $K$, $Q$, $V$ represents the key, value, and query features respectively.

\subsection{Personalized Inpainting Diffusion Model Adaptation}
\label{sec:personalized_inpaiting}

\begin{wrapfigure}{r}{0.5\textwidth}
  \centering
   \includegraphics[width=1\linewidth]{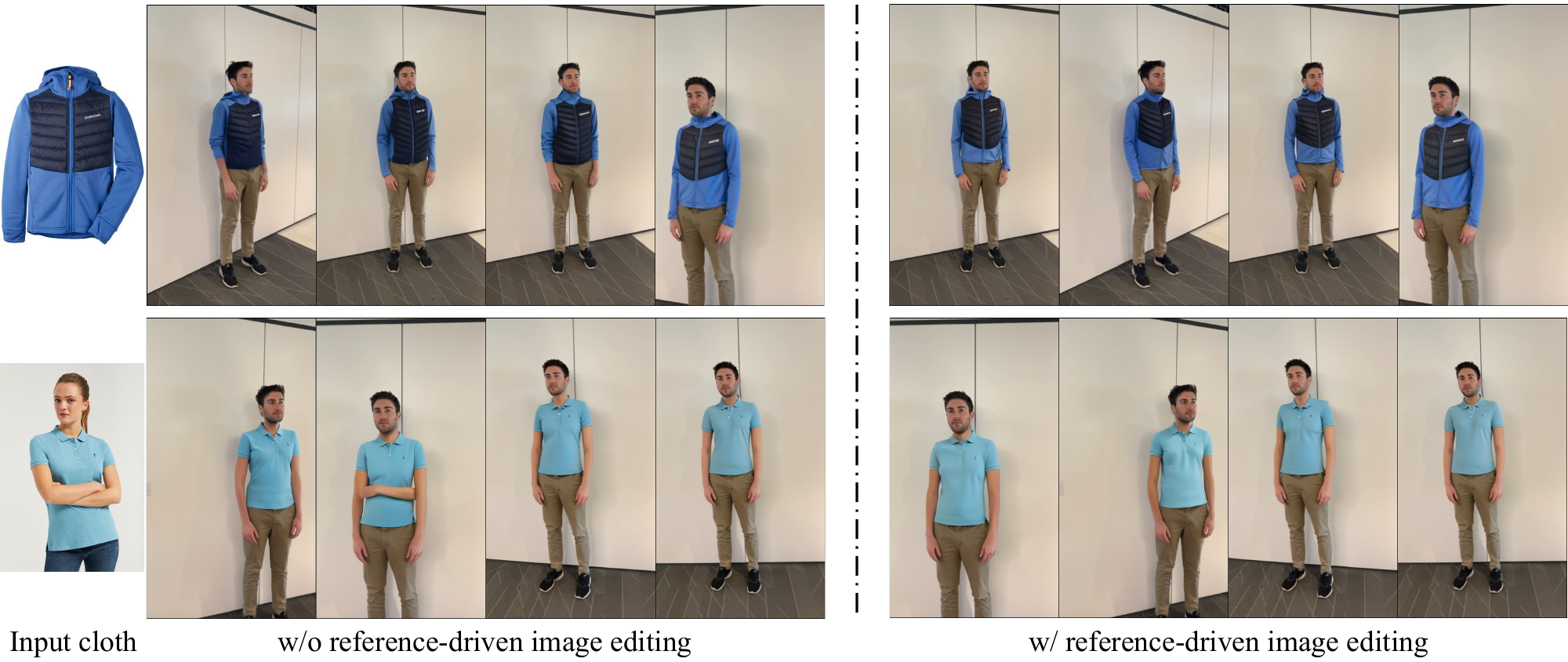}   
  \caption{\textbf{Effectiveness of reference-driven image editing in multi-view image editing.} }
   \label{fig:att1-motivation}
   \vspace{-1em}
\end{wrapfigure}

Existing methods for editing 3D scenes~\citep{haque2023instructnerf2nerf, ig2g, wu2024gaussctrl, dong2024vica, zhuang2024tip} typically rely on a pre-trained diffusion model to control the editing process and update the training dataset. However, these approaches would struggle with tasks such as modifying the garment of a human subject (see Fig.~\ref{fig:comparison}). A notable cause is that diffusion models like instruct-pix2pix~\citep{brooks2022instructpix2pix} lack the capability to accurately perceive and edit clothing locally. Although there have been advancements in diffusion models~\citep{choi2024improving, zeng2024cat, zhu2023tryondiffusion} for 2D virtual try-on, applying them directly to 3D scene editing often leads to inconsistencies and geometric distortions. This is primarily due to the inherent randomness of diffusion models, which struggle to accurately predict how garments will appear from different viewpoints, leading to discrepancies across various views (see Fig.~\ref{fig:att1-motivation}).
To tackle this problem in 3D virtual try-on, we propose injecting spatial consistent features derived from the training dataset $\mathcal{I}_{\text{train}}$ into the diffusion model.

\paragraph{Personalized Diffusion Model via LoRA fine-tuning.} 
Low-Rank Adaption (LoRA)~\citep{hu2021lora} is a technique designed to efficiently fine-tune large language models, and has recently been extended to diffusion models. Rather than adjusting the entire model, LoRA focuses on modifying a low-rank residual component $\Delta \theta$, which is represented as a sum of low-rank matrices. 
This method allows us to incorporate characteristics of a specific image into the learned distribution of a pre-trained diffusion model.

In order to design an image-prompted network, we first apply LoRA to enhance a pre-trained Stable Diffusion Inpainting Model~\citep{rombach2022high}. 
Specifically, it involves training the LoRA component $\Delta \theta$ using a collection of edited training images $X_{\text{train}} = \{I_i | i \in [0, n)\}$, where $n$ represents the total number of images, with the following objective:
\begin{equation}
    \mathcal{L}(\Delta\theta) = \mathbb{E}_{\epsilon, t}[||\epsilon - \epsilon_{\theta + \Delta \theta} (\sqrt{a_t}\mathbf{z}_{0-i} + \sqrt{1 - a_t}\epsilon, t, y)||^2],
\end{equation}
where $\mathbf{z}_0 = \mathcal{E}(I_i)$ is the latent embedding from the VAE encoder for image $I_i$, $\epsilon$ is the randomly sampled Gaussian noise, $y$ denotes the text embedding, and $\epsilon_{\theta + \Delta \theta}$ represents the UNet model enhanced with LoRA.

To further enhance the performance, we generate $K$ random binary masks $\mathcal{M} = \{m_i = 0, 1 | i\in[0, K)\}$ and apply these masks to the images~\citep{tang2024realfill} during LoRA fine-tuning. Then the objective becomes:
\begin{equation}
    \mathcal{L}(\Delta\theta_i) = \mathbb{E}_{\epsilon, t}[||\epsilon - \epsilon_{\theta + \Delta \theta} (\sqrt{a_t}\mathbf{z}_{0-i} \odot (1 - m_i) + \sqrt{1 - a_t}\epsilon, t, y)||^2],
\end{equation}
where $\odot$ denotes the element-wise product.

\paragraph{Reference-driven Image Editing.} 
To achieve a well-trained LoRA model, the first critical step is constructing the edited training image set $X_{\text{train}}$. To this end, we further propose reference-driven image editing. Na\"ively, one might consider such a straightforward method: applying images from the input human images $\mathcal{I}_{\text{train}}$ directly to a pre-trained 2D virtual try-on diffusion model to obtain the edited images individually. However, we found that this method introduces significant inconsistencies in garment appearance, which adversely affects the quality and reliability of the LoRA model, as shown in Fig.~\ref{fig:att1-motivation}. We attribute this problem to the randomness of the Gaussian noise, which would lead to variations in the attention features. 

Drawing inspiration from recent advancements in temporal-aware self-attention techniques used in video generation~\citep{zhou2024storydiffusion, chen2023videocrafter1, chen2024videocrafter2, blattmann2023stable}, we propose a novel approach to enhance image consistency using a pre-trained IDM-VTON~\citep{choi2024improving}. Our approach involves first creating an image set $X_{\text{train}}$ through random sampling of $n$ images from the input multi-view human images $\mathcal{I}_{\text{train}}$. Note that we set $n = 4$ for the experiments reported in this paper. We then perform simultaneous editing of these images while incorporating reference attention features into the denoising process to enhance the overall consistency.
Specifically, during the denoising step $t$, we begin by processing the latent features $\mathbf{z}_{t-i}$ of the images $I_i \in X_{\text{train}}$ through the UNet of IDM-VTON, which produces the key and value matrices $K_{t-i}$ and $V_{t-i}$ for the self-attention mechanism. We then integrate reference attention features to update these matrices accordingly:
\begin{equation}
    K_{t-i} := [K_{t-i}, K_{t-\text{ref}}], \quad V_{t-i} := [V_{t-i}, V_{t-\text{ref}}], \quad i = 0, ..., n
\end{equation}
where $[\cdot]$ represents the concatenation operation. In our implementation, we treat the first image as the reference image, \ie, $K_{t-\text{ref}} = K_{t-0}$, $V_{t-\text{ref}} = V_{t-0}$. We then replace the corresponding matrices in the UNet with these updated values to obtain the edited images:
\begin{equation}
    X_{\text{train}} := \{F_{\theta}(I_i, I_{\text{ref}}) | i = 0, ..., n-1\}
\end{equation}
where $F_{\theta}(\cdot)$ denotes the pre-trained IDM-VTON model. 
This approach ensures that during the denoising steps, the intermediate latents are influenced by consistent reference features, thereby improving the overall consistency of the edited images.

\subsection{Persona-aware 3DGS Editing}
\label{sec:3dgs_editing}
After developing a fine-tuned personalized inpainting diffusion model, integrating it into the 3DGS editing pipeline introduces additional challenges. Unfortunately, images generated by this fine-tuned diffusion model can still exhibit inconsistencies, particularly when the rendered viewpoints differ significantly from those in the edited image set $X_{\text{train}}$. Consequently, this can negatively impact 3DGS editing by introducing visual artifacts and inconsistent textures (see Fig.~\ref{fig:ablation-wo-persona}). The problem stems from the limited number of training images used during fine-tuning, which restricts the model’s ability to produce consistent features across various viewpoints. This issue remains even when we increase the number of images for LoRA fine-tuning (see Appx.~\ref{sec:number-views}), which also raises GPU memory requirements and reduces training efficiency.

To address this, we propose persona-aware 3DGS editing, which refines diffusion process by merging two predicted attention features: one based on the editing direction and the other derived from the edited image set $X_{\text{train}}$:
\begin{equation}
    \mathtt{ATT}(Q_j, K_j, V_j) := \lambda \cdot \mathtt{ATT}(Q_j, K_j, V_j) + (1 - \lambda) \cdot \frac{1}{n} \sum_{i\in {X_{\text{train}}}} \mathtt{ATT}(Q_j, K_i, V_i),
\end{equation}
where $\lambda$ is a hyper-parameter to balance the effects, and defaults to 0.55 in our experiments. Instead of adapting the original stable diffusion inpainting model with LoRA, we adapt it via a ControlNet-based stable diffusion inpainting model to condition the inpainting process on the input garment image, thus enhancing the fidelity of the results. Formally, given a rendered image $I_\text{src}$ from 3DGS scene and a garment image $I_{\text{cloth}}$ with captioning text $y$ from BLIP-2~\citep{li2023blip}, we first input these into the fine-tuned personalized inpainting diffusion model equipped with ControlNet $\mathcal{C}$ to obtain the edited image:
\begin{equation}
    I_{\text{edit}} = \epsilon_{\theta + \Delta \theta}(\mathbf{z}_{\text{src}}; y, t, \mathcal{C}(I_{\text{cloth}})),
\end{equation}
where $\mathbf{z}_{\text{src}}$ represents the encoded latents from the rendered image. Our optimization objective is then be formulated as:
\begin{equation}
    \mathcal{L} = \lambda_1 \cdot \mathcal{L}_{\text{MAE}}(I_{\text{edit}}, I_{\text{src}}) + \lambda_2 \cdot \mathcal{L}_{\text{LPIPS}}(I_{\text{edit}}, I_{\text{src}})
\end{equation}
where $\lambda_1$ and $\lambda_2$ are hyper-parameters, which defaults to $10$ and $15$ respectively.

\begin{figure}[t]
  \centering
   \includegraphics[width=1\linewidth]{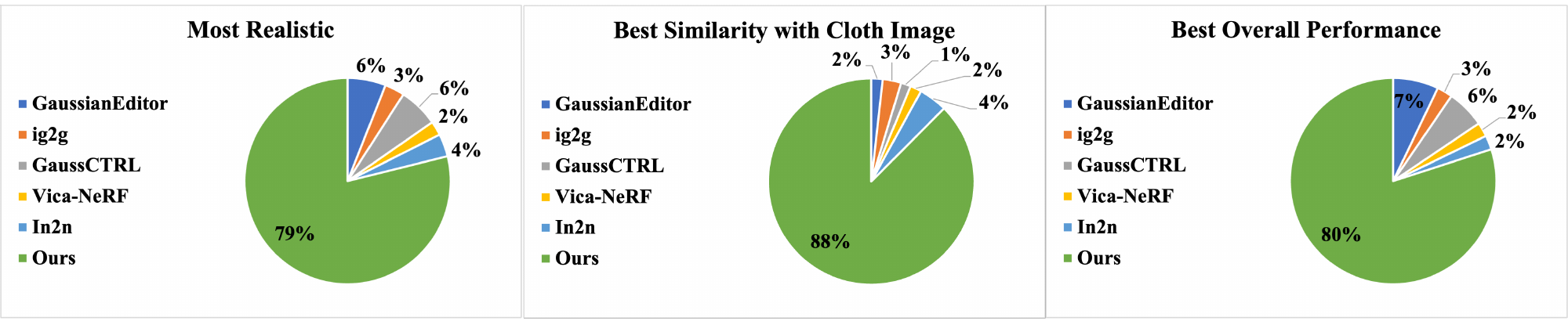}   
   \vspace{-1.5em}
  \caption{\textbf{User study.}  Numbers are averaged over 625 responses from 25 volunteers.}
   \label{fig:user-studies}
   \vspace{-1em}
\end{figure} 
\subsection{Implementation Details}
\OMO builds upon official implementation of GaussianEditor~\citep{chen2023gaussianeditor} for 3DGS editing. 
While GaussianEditor uses a large language model~\citep{kirillov2023segment} to create a 2D image mask and then invert it for labeling locally edited 3D Gaussians, we take a different approach by employing a 2D human parsing model~\citep{li2020self} and a human pose estimation model~\citep{Güler_2018_CVPR} to generate the image mask. 
For our personalized inpainting diffusion model, we utilize the Stable-Diffusion-2-Inpainting model~\citep{stable-inpaint-diffusion} and adopt hyperparameters from RealFill~\citep{tang2024realfill}. 
We utilize the pre-trained BLIP-2 model to generate captions for the garment image, which serves as part of the input to the diffusion model.
Unlike many existing 3D editing methods that are limited to a maximum image resolution of $512\times512$ due to constraints from Instruct-pix2pix, \OMO can operate without such limitations, allowing edits at the original resolution of the 3D scene. 
Additionally, while other methods may adjust hyperparameters for different scenes, we keep all hyperparameters fixed across our experiments. 
For experiments reported in this paper, we fine-tune the LoRA module for 1,000 iterations, while the 3DGS editing stage involves 4,000 iterations. Typically, the fine-tuning of the LoRA module takes about 30 minutes, and the 3DGS editing requires approximately 25 minutes on a single V100 GPU with 32GB of memory.

\section{Experiments}
\label{sec:experiment}
We now evaluate the performance of our \OMO both quantitatively and qualitatively, and provide comparisons with other SOTA methods for 3D scene editing.

\paragraph{3D-VTONBench.}
Existing virtual try-on techniques primarily focus on 2D image generation, while the majority of 3D virtual try-on methods~\citep{rong2024gaussian, feng2022capturing, jiang2020bcnet, corona2021smplicit, pons2017clothcap, grigorev2023hood} are centered around dressing the SMPL models~\citep{SMPL:2015, SMPL-X:2019} with human garments. On the other hand, current 3D scene editing approaches tend to work with general scenes, leaving 3D virtual try-on underexplored. As a result, there is a notable lack of specific evaluation benchmarks for this task. To thoroughly assess the effectiveness of our methods, we introduce 3D-VTONBench, the first benchmark dataset dedicated to evaluating 3D virtual try-on. Our dataset includes 60 data subjects captured in various poses and garments. We believe that 3D-VTONBench will foster further research in this important area. 

\paragraph{Comparison Methods.}
We compare the editing results with five techniques: GaussianEditor~\citep{chen2023gaussianeditor}, Instruct-Gaussian2Gaussian (IG2G)~\citep{ig2g}, GaussCTRL~\citep{wu2024gaussctrl}, Instruct-NeRF2NeRF (IN2N)~\citep{haque2023instructnerf2nerf}, and Vica-NeRF~\citep{dong2024vica}. Since these methods only accept text prompts as input, we use ChatGPT to generate the text prompts corresponding to the clothing images. We don't compare with GaussianVTON~\citep{chen2024gaussianvton} as their code is not publicly available.

\begin{figure}[t]
  \centering
   \includegraphics[width=\linewidth]{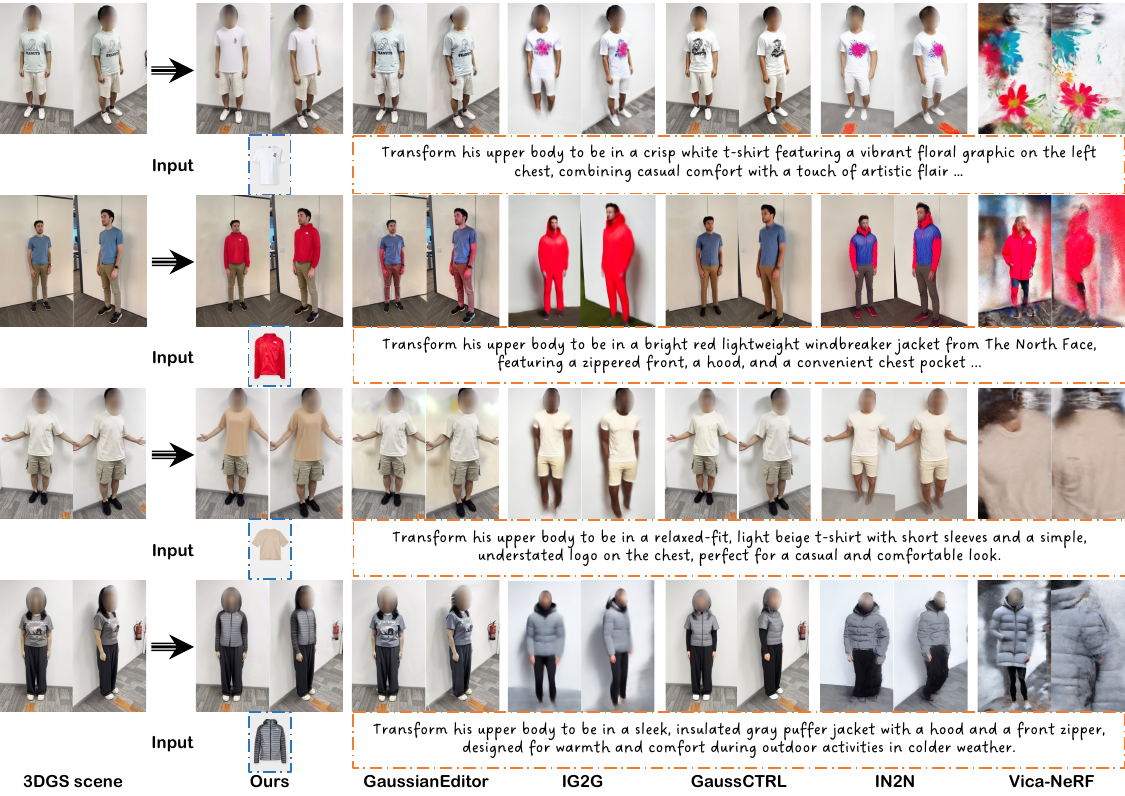}
  \caption{\textbf{Qualitative comparison with existing 3D scene editing techniques.} In contrast to other methods that often struggle to produce satisfactory virtual try-on results, our approach consistently delivers high-quality geometry and texture, closely resembling the input garment image.}
   \label{fig:comparison}
\end{figure} 
\subsection{Quantitative Evaluations}
\paragraph{User Studies.}
We begin by conducting a series of user studies with 25 pairs of edited results to assess the quality of our method. For each pair, we presented the videos generated by our method alongside those from five comparison methods~\citep{chen2023gaussianeditor, ig2g, haque2023instructnerf2nerf, dong2024vica, wu2024gaussctrl}. Participants were asked to watch these videos and select the best result based on (1) realism, (2) similarity to the clothing image, and (3) overall performance. 
A total of 25 volunteers participated in the user studies, providing 625 responses overall. The results, provided in Fig.~\ref{fig:user-studies}, show that our method significantly outperformed the others across all three dimensions. Furthermore, the evaluation of similarity to the clothing image highlights the limitations of text descriptions in conveying garment details, emphasizing the necessity for our image-prompted pipeline.

\subsection{Qualitative Evaluations}
\label{sec:qualitative}

\begin{wrapfigure}{r}{0.5\textwidth}
  \centering
   \includegraphics[width=1\linewidth]{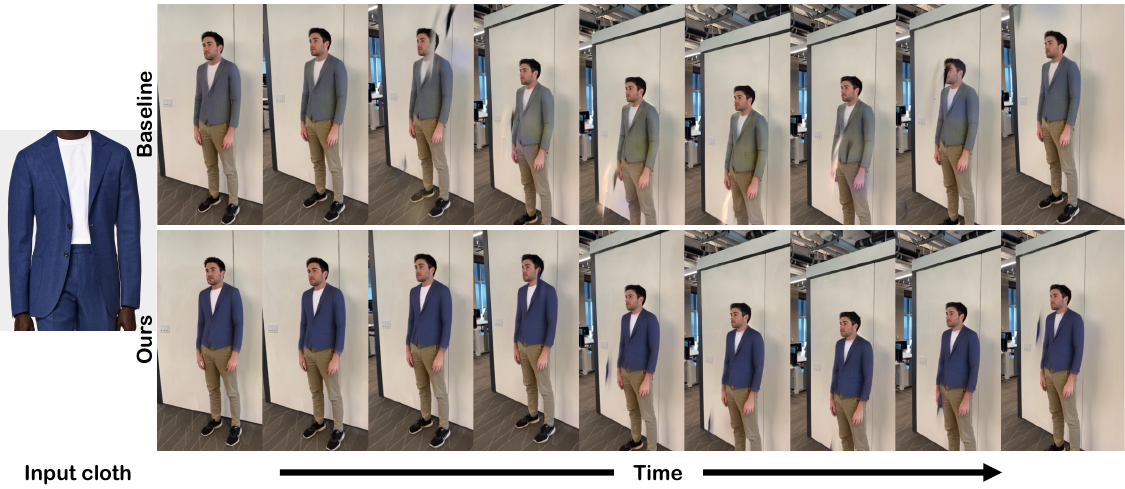}   
   \vspace{-2em}
  \caption{\textbf{Comparison with baseline method.} }
   \label{fig:ablation-baseline}
   \vspace{-1em}
\end{wrapfigure}
\paragraph{Comparison with baseline method.}
We begin with qualitative evaluations to first compare our approach against the baseline method. Specifically, the baseline method achieves 3D virtual try-on by (1) generating edited training image set $X_{\text{train}}$ individually via IDM-VTON~\citep{choi2024improving}; (2) fine-tuning LoRA module; (3) editing the 3D scene with fine-tuned model. 
Results are provided in Fig.~\ref{fig:ablation-baseline}.
The results reveal that the baseline method encounters challenges in three main areas of 3D virtual try-on:
\textbf{(1)} it has trouble generating outputs that closely resemble the input garment image; 
\textbf{(2)} it struggles to maintain consistency across different frames; 
and \textbf{(3)} it tends to produce artifacts, such as outliers. 
In contrast, our contributions, which include reference-driven image editing and persona-aware 3DGS editing, effectively lead to consistent results that align closely with the garment image.

\paragraph{Comparisons with SOTA methods.}
We provide visual comparisons with existing methods in Fig.~\ref{fig:comparison}, from which we can draw the following conclusions: 
\textbf{(1)} Textual prompts, even when carefully refined, often struggle to capture the details of garments. This limitation contributes to the tendency of existing methods to produce suboptimal 3D scenes for virtual try-on compared to our approach;
\textbf{(2)} While GaussianEditor~\citep{chen2023gaussianeditor} enables local editing using a large language model~\citep{kirillov2023segment}, it has difficulty making substantial changes to the original geometry and textures. This leads to 3D scenes that do not accurately reflect the textual descriptions;
\textbf{(3)} GaussCTRL~\citep{wu2024gaussctrl} utilizes a depth-conditioned ControlNet~\citep{zhang2023controlnet} to tackle inconsistency issues. However, it struggles with (i) preserving the original identity and (ii) producing results with insufficient editing;
\textbf{(4)} Instruct-NeRF2NeRF~\citep{haque2023instructnerf2nerf} and Instruct-Gaussian2Gaussian~\citep{ig2g} effectively extract information from text inputs, yet they struggle to (i) keep the background unchanged, (ii) maintain the original identity and poses, and (iii) produce high-resolution renderings;
\textbf{(5)} Although Vica-NeRF~\citep{dong2024vica} performs well with general scenes, it has difficulty editing human-centric 3D environments.
In contrast, our method consistently produces superior results, offering higher-quality details in both geometry and texture, along with strong consistency with the provided garment image. 
Additional comparisons can be found in the Appendix.

\subsection{Ablation Study}
\label{sec:ablation}


\begin{wrapfigure}{r}{0.5\textwidth}
  \centering
   \includegraphics[width=1\linewidth]{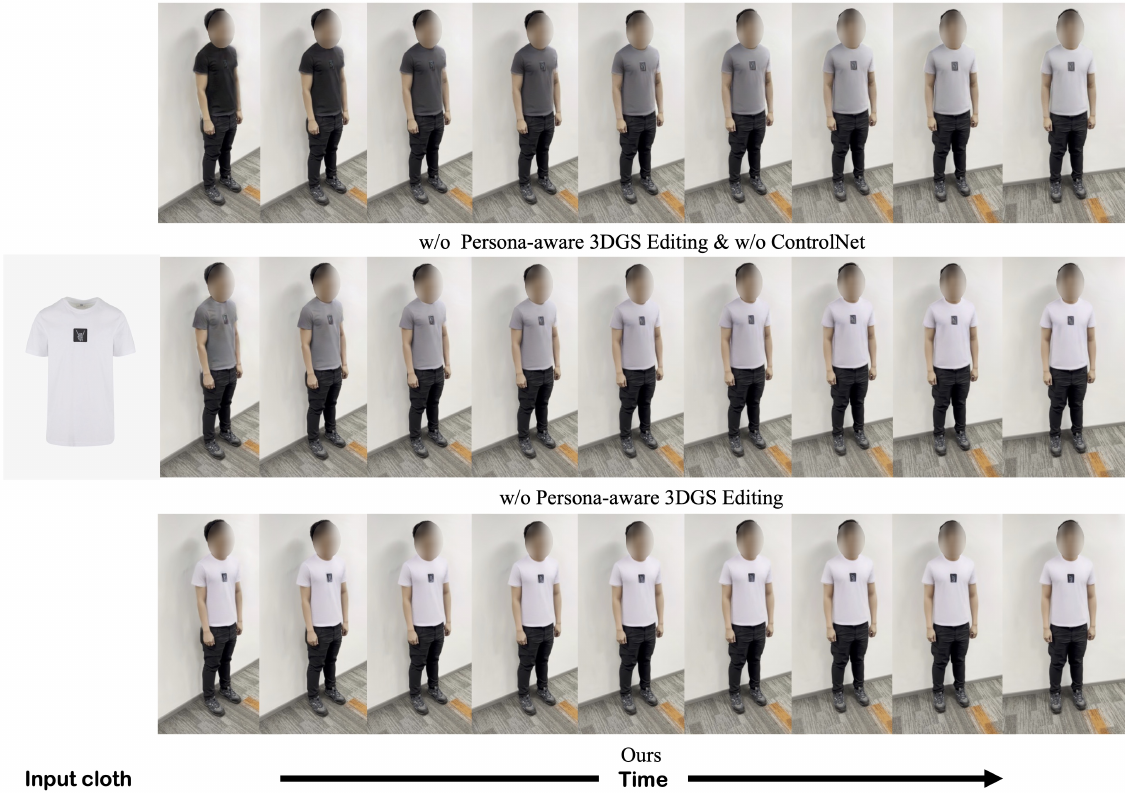}   
  \caption{\textbf{Analysis of persona-aware 3DGS editing and the utilization of ControlNet.} }
   \label{fig:ablation-wo-persona}
   \vspace{-1em}
\end{wrapfigure}
\paragraph{Effectiveness of Persona-aware 3DGS Editing.}
We then conduct ablation studies to assess our persona-aware 3DGS editing and the use of ControlNet, with results shown in Fig.~\ref{fig:ablation-wo-persona}. Both components are essential for ensuring consistent 3D scene editing; without them, the edited scenes struggle to (1) maintain consistent texture across frames and (2) match the texture of the input garment.

\begin{wrapfigure}{r}{0.5\textwidth}
  \centering
   \includegraphics[width=1\linewidth]{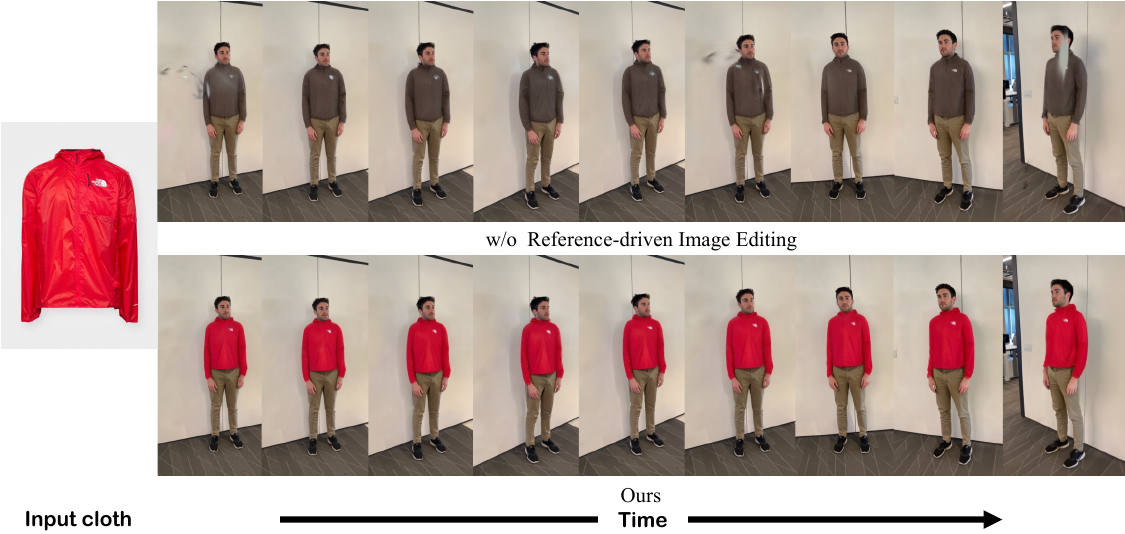}   
  \caption{\textbf{Effectiveness of reference-driven image editing for 3D virtual try-on.} }
   \label{fig:ablation-woatt1}
   \vspace{-1em}
\end{wrapfigure}
\paragraph{Effectiveness of Reference-driven Image Editing.}
In Fig.~\ref{fig:ablation-woatt1}, we present ablation studies to assess the effect of our proposed reference-driven image editing.
Existing diffusion models for 2D virtual try-on often demonstrate inconsistencies when editing multi-view images individually (as shown in Fig.\ref{fig:att1-motivation}). 
This inconsistency can hinder the effective fine-tuning of the LoRA module, resulting in subpar 3DGS editing. 
For instance, the results shown in Fig.~\ref{fig:ablation-woatt1}, edited without our design, display a mismatch in texture with the input garment image. 
In contrast, our reference-driven image editing effectively addresses this issue, yielding high-fidelity 3D edits with textures that remain consistent with the input.

\section{Conclusion}
In this paper, we have introduced \OM, a novel image-prompted method for 3D virtual try-on. 
We first propose a personalized diffusion adaptation through LoRA fine-tuning, allowing the model to better represent the input garment by extending its pre-trained distribution. 
Additionally, we introduce reference-driven image editing to enable consistent multi-view editing, providing a solid foundation for LoRA fine-tuning. 
To further enhance multi-view consistency in the edited 3D scenes, we present persona-aware 3DGS editing, which respects both the desired editing direction and features derived from the original edited images.
Extensive evaluations demonstrate the effectiveness of our design, highlighting that \OMO delivers high-fidelity results across a range of scenarios and significantly outperforms state-of-the-art methods.

\paragraph{Limitations.}
While establishing a new state-of-the-art for 3D virtual try-on, our \OMO approach still has some limitations: (1) Inheriting biases from pre-trained 2D virtual try-on models, our pipeline has difficulty accurately modeling long hair when it intersects with clothing. (2) Although our method can accommodate human subjects in various poses, it encounters challenges with severe self-occlusion, such as when a person crosses their arms in front of the chest.

\paragraph{Ackowledgements.}
Huge thanks go to Haozhe Xie and Jiuyun Zhang for their precious help in collecting the datasets. Authors marked with ``equal contribution'' are allowed to switch their orders in the author list in their resumes and websites.

\bibliography{reb}

\begin{thebibliography}{100}
\providecommand{\natexlab}[1]{#1}
\providecommand{\url}[1]{\texttt{#1}}
\expandafter\ifx\csname urlstyle\endcsname\relax
  \providecommand{\doi}[1]{doi: #1}\else
  \providecommand{\doi}{doi: \begingroup \urlstyle{rm}\Url}\fi

\bibitem[Balaji et~al.(2022)Balaji, Nah, Huang, Vahdat, Song, Kreis, Aittala, Aila, Laine, Catanzaro, et~al.]{balaji2022ediffi}
Yogesh Balaji, Seungjun Nah, Xun Huang, Arash Vahdat, Jiaming Song, Karsten Kreis, Miika Aittala, Timo Aila, Samuli Laine, Bryan Catanzaro, et~al.
\newblock ediffi: Text-to-image diffusion models with an ensemble of expert denoisers.
\newblock \emph{arXiv preprint arXiv:2211.01324}, 2022.

\bibitem[Blattmann et~al.(2023{\natexlab{a}})Blattmann, Dockhorn, Kulal, Mendelevitch, Kilian, Lorenz, Levi, English, Voleti, Letts, et~al.]{blattmann2023stable}
Andreas Blattmann, Tim Dockhorn, Sumith Kulal, Daniel Mendelevitch, Maciej Kilian, Dominik Lorenz, Yam Levi, Zion English, Vikram Voleti, Adam Letts, et~al.
\newblock Stable video diffusion: Scaling latent video diffusion models to large datasets.
\newblock \emph{arXiv preprint arXiv:2311.15127}, 2023{\natexlab{a}}.

\bibitem[Blattmann et~al.(2023{\natexlab{b}})Blattmann, Rombach, Ling, Dockhorn, Kim, Fidler, and Kreis]{blattmann2023align}
Andreas Blattmann, Robin Rombach, Huan Ling, Tim Dockhorn, Seung~Wook Kim, Sanja Fidler, and Karsten Kreis.
\newblock Align your latents: High-resolution video synthesis with latent diffusion models.
\newblock In \emph{CVPR}, 2023{\natexlab{b}}.

\bibitem[Blattmann et~al.(2023{\natexlab{c}})Blattmann, Rombach, Ling, Dockhorn, Kim, Fidler, and Kreis]{blattmann2023videoldm}
Andreas Blattmann, Robin Rombach, Huan Ling, Tim Dockhorn, Seung~Wook Kim, Sanja Fidler, and Karsten Kreis.
\newblock Align your latents: High-resolution video synthesis with latent diffusion models.
\newblock In \emph{CVPR}, 2023{\natexlab{c}}.

\bibitem[Brooks et~al.(2023)Brooks, Holynski, and Efros]{brooks2022instructpix2pix}
Tim Brooks, Aleksander Holynski, and Alexei~A Efros.
\newblock Instructpix2pix: Learning to follow image editing instructions.
\newblock In \emph{IEEE Conference on Computer Vision and Pattern Recognition}, 2023.

\bibitem[Canny(1986)]{canny1986computational}
John Canny.
\newblock A computational approach to edge detection.
\newblock \emph{IEEE Transactions on Pattern Analysis and Machine Intelligence}, 1986.

\bibitem[Chen et~al.(2024{\natexlab{a}})Chen, Huang, Huang, Ge, and Shao]{chen2024gaussianvton}
Haodong Chen, Yongle Huang, Haojian Huang, Xiangsheng Ge, and Dian Shao.
\newblock Gaussianvton: 3d human virtual try-on via multi-stage gaussian splatting editing with image prompting.
\newblock \emph{arXiv preprint arXiv:2405.07472}, 2024{\natexlab{a}}.

\bibitem[Chen et~al.(2023{\natexlab{a}})Chen, Xia, He, Zhang, Cun, Yang, Xing, Liu, Chen, Wang, et~al.]{chen2023videocrafter1}
Haoxin Chen, Menghan Xia, Yingqing He, Yong Zhang, Xiaodong Cun, Shaoshu Yang, Jinbo Xing, Yaofang Liu, Qifeng Chen, Xintao Wang, et~al.
\newblock Videocrafter1: Open diffusion models for high-quality video generation.
\newblock \emph{arXiv preprint arXiv:2310.19512}, 2023{\natexlab{a}}.

\bibitem[Chen et~al.(2024{\natexlab{b}})Chen, Zhang, Cun, Xia, Wang, Weng, and Shan]{chen2024videocrafter2}
Haoxin Chen, Yong Zhang, Xiaodong Cun, Menghan Xia, Xintao Wang, Chao Weng, and Ying Shan.
\newblock Videocrafter2: Overcoming data limitations for high-quality video diffusion models.
\newblock In \emph{Proceedings of the IEEE/CVF Conference on Computer Vision and Pattern Recognition}, pp.\  7310--7320, 2024{\natexlab{b}}.

\bibitem[Chen et~al.(2023{\natexlab{b}})Chen, Chen, Zhang, Wang, Yang, Wang, Cai, Yang, Liu, and Lin]{chen2023gaussianeditor}
Yiwen Chen, Zilong Chen, Chi Zhang, Feng Wang, Xiaofeng Yang, Yikai Wang, Zhongang Cai, Lei Yang, Huaping Liu, and Guosheng Lin.
\newblock Gaussianeditor: Swift and controllable 3d editing with gaussian splatting.
\newblock \emph{arXiv preprint arXiv:2311.14521}, 2023{\natexlab{b}}.

\bibitem[Cheng et~al.(2023)Cheng, Yang, Wang, Li, Zhang, Zhang, and Yuan]{cheng2023progressive3d}
Xinhua Cheng, Tianyu Yang, Jianan Wang, Yu~Li, Lei Zhang, Jian Zhang, and Li~Yuan.
\newblock Progressive3d: Progressively local editing for text-to-3d content creation with complex semantic prompts.
\newblock \emph{arXiv preprint arXiv:2310.11784}, 2023.

\bibitem[Choi et~al.(2021)Choi, Park, Lee, and Choo]{choi2021viton}
Seunghwan Choi, Sunghyun Park, Minsoo Lee, and Jaegul Choo.
\newblock Viton-hd: High-resolution virtual try-on via misalignment-aware normalization.
\newblock In \emph{Proceedings of the IEEE/CVF conference on computer vision and pattern recognition}, pp.\  14131--14140, 2021.

\bibitem[Choi et~al.(2024)Choi, Kwak, Lee, Choi, and Shin]{choi2024improving}
Yisol Choi, Sangkyung Kwak, Kyungmin Lee, Hyungwon Choi, and Jinwoo Shin.
\newblock Improving diffusion models for virtual try-on.
\newblock \emph{arXiv preprint arXiv:2403.05139}, 2024.

\bibitem[Corona et~al.(2021)Corona, Pumarola, Alenya, Pons-Moll, and Moreno-Noguer]{corona2021smplicit}
Enric Corona, Albert Pumarola, Guillem Alenya, Gerard Pons-Moll, and Francesc Moreno-Noguer.
\newblock Smplicit: Topology-aware generative model for clothed people.
\newblock In \emph{Proceedings of the IEEE/CVF conference on computer vision and pattern recognition}, pp.\  11875--11885, 2021.

\bibitem[Cyrus \& Ayyan(2023)Cyrus and Ayyan]{ig2g}
Vachha Cyrus and Haque Ayyan.
\newblock Instruct-gaussian2gaussian: Editing 3d gaussian splatting scenes with instructions.
\newblock \url{https://instruct-gs2gs.github.io/}, 2023.

\bibitem[Dhariwal \& Nichol(2020)Dhariwal and Nichol]{dhariwal2021beatgan}
Prafulla Dhariwal and Alexander Nichol.
\newblock Diffusion models beat gans on image synthesis.
\newblock In \emph{Advances in Neural Information Processing Systems}, 2020.

\bibitem[Dong \& Wang(2024)Dong and Wang]{dong2024vica}
Jiahua Dong and Yu-Xiong Wang.
\newblock Vica-nerf: View-consistency-aware 3d editing of neural radiance fields.
\newblock \emph{Advances in Neural Information Processing Systems}, 36, 2024.

\bibitem[Esser et~al.(2023)Esser, Chiu, Atighehchian, Granskog, and Germanidis]{esser2023gen-1}
Patrick Esser, Johnathan Chiu, Parmida Atighehchian, Jonathan Granskog, and Anastasis Germanidis.
\newblock Structure and content-guided video synthesis with diffusion models.
\newblock \emph{arXiv preprint arXiv:2302.03011}, 2023.

\bibitem[Feng et~al.(2022)Feng, Yang, Pollefeys, Black, and Bolkart]{feng2022capturing}
Yao Feng, Jinlong Yang, Marc Pollefeys, Michael~J Black, and Timo Bolkart.
\newblock Capturing and animation of body and clothing from monocular video.
\newblock In \emph{SIGGRAPH Asia 2022 Conference Papers}, pp.\  1--9, 2022.

\bibitem[Gal et~al.(2023)Gal, Alaluf, Atzmon, Patashnik, Bermano, Chechik, and Cohen-Or]{gal2022textual-inversion}
Rinon Gal, Yuval Alaluf, Yuval Atzmon, Or~Patashnik, Amit~H Bermano, Gal Chechik, and Daniel Cohen-Or.
\newblock An image is worth one word: Personalizing text-to-image generation using textual inversion.
\newblock In \emph{ICLR}, 2023.

\bibitem[Ge et~al.(2021{\natexlab{a}})Ge, Song, Ge, Yang, Liu, and Luo]{ge2021disentangled}
Chongjian Ge, Yibing Song, Yuying Ge, Han Yang, Wei Liu, and Ping Luo.
\newblock Disentangled cycle consistency for highly-realistic virtual try-on.
\newblock In \emph{Proceedings of the IEEE/CVF conference on computer vision and pattern recognition}, pp.\  16928--16937, 2021{\natexlab{a}}.

\bibitem[Ge et~al.(2021{\natexlab{b}})Ge, Song, Zhang, Ge, Liu, and Luo]{ge2021parser}
Yuying Ge, Yibing Song, Ruimao Zhang, Chongjian Ge, Wei Liu, and Ping Luo.
\newblock Parser-free virtual try-on via distilling appearance flows.
\newblock In \emph{Proceedings of the IEEE/CVF conference on computer vision and pattern recognition}, pp.\  8485--8493, 2021{\natexlab{b}}.

\bibitem[Goodfellow et~al.(2014)Goodfellow, Pouget-Abadie, Mirza, Xu, Warde-Farley, Ozair, Courville, and Bengio]{goodfellow2014generative}
Ian Goodfellow, Jean Pouget-Abadie, Mehdi Mirza, Bing Xu, David Warde-Farley, Sherjil Ozair, Aaron Courville, and Yoshua Bengio.
\newblock Generative adversarial nets.
\newblock \emph{Advances in neural information processing systems}, 27, 2014.

\bibitem[Gou et~al.(2023)Gou, Sun, Zhang, Si, Qian, and Zhang]{gou2023taming}
Junhong Gou, Siyu Sun, Jianfu Zhang, Jianlou Si, Chen Qian, and Liqing Zhang.
\newblock Taming the power of diffusion models for high-quality virtual try-on with appearance flow.
\newblock In \emph{Proceedings of the 31st ACM International Conference on Multimedia}, pp.\  7599--7607, 2023.

\bibitem[Grigorev et~al.(2023)Grigorev, Black, and Hilliges]{grigorev2023hood}
Artur Grigorev, Michael~J Black, and Otmar Hilliges.
\newblock Hood: Hierarchical graphs for generalized modelling of clothing dynamics.
\newblock In \emph{Proceedings of the IEEE/CVF Conference on Computer Vision and Pattern Recognition}, pp.\  16965--16974, 2023.

\bibitem[Guo et~al.(2023)Guo, Zheng, Hou, Gao, Deng, Ma, Hu, Zha, Huang, Wan, et~al.]{guo2023i2v}
Xun Guo, Mingwu Zheng, Liang Hou, Yuan Gao, Yufan Deng, Chongyang Ma, Weiming Hu, Zhengjun Zha, Haibin Huang, Pengfei Wan, et~al.
\newblock I2v-adapter: A general image-to-video adapter for video diffusion models.
\newblock \emph{arXiv preprint arXiv:2312.16693}, 2023.

\bibitem[Güler et~al.(2018)Güler, Neverova, and Kokkinos]{Güler_2018_CVPR}
Rıza~Alp Güler, Natalia Neverova, and Iasonas Kokkinos.
\newblock Densepose: Dense human pose estimation in the wild.
\newblock In \emph{Proceedings of the IEEE Conference on Computer Vision and Pattern Recognition (CVPR)}, June 2018.

\bibitem[Han et~al.(2023)Han, Cao, Han, Zhu, Deng, Song, Xiang, and Wong]{han2023headsculpt}
Xiao Han, Yukang Cao, Kai Han, Xiatian Zhu, Jiankang Deng, Yi-Zhe Song, Tao Xiang, and Kwan-Yee~K Wong.
\newblock Headsculpt: Crafting 3d head avatars with text.
\newblock \emph{arXiv preprint arXiv:2306.03038}, 2023.

\bibitem[Han et~al.(2018)Han, Wu, Wu, Yu, and Davis]{han2018viton}
Xintong Han, Zuxuan Wu, Zhe Wu, Ruichi Yu, and Larry~S Davis.
\newblock Viton: An image-based virtual try-on network.
\newblock In \emph{Proceedings of the IEEE conference on computer vision and pattern recognition}, pp.\  7543--7552, 2018.

\bibitem[Haque et~al.(2023)Haque, Tancik, Efros, Holynski, and Kanazawa]{haque2023instructnerf2nerf}
Ayaan Haque, Matthew Tancik, Alexei~A Efros, Aleksander Holynski, and Angjoo Kanazawa.
\newblock Instruct-nerf2nerf: Editing 3d scenes with instructions.
\newblock \emph{arXiv preprint arXiv:2303.12789}, 2023.

\bibitem[Hauswiesner et~al.(2013)Hauswiesner, Straka, and Reitmayr]{hauswiesner2013virtual}
Stefan Hauswiesner, Matthias Straka, and Gerhard Reitmayr.
\newblock Virtual try-on through image-based rendering.
\newblock \emph{IEEE transactions on visualization and computer graphics}, 19\penalty0 (9):\penalty0 1552--1565, 2013.

\bibitem[He et~al.(2016)He, Zhang, Ren, and Sun]{he2016deep}
Kaiming He, Xiangyu Zhang, Shaoqing Ren, and Jian Sun.
\newblock Deep residual learning for image recognition.
\newblock In \emph{IEEE Conference on Computer Vision and Pattern Recognition}, 2016.

\bibitem[Hertz et~al.(2022)Hertz, Mokady, Tenenbaum, Aberman, Pritch, and Cohen-Or]{hertz2022prompt2prompt}
Amir Hertz, Ron Mokady, Jay Tenenbaum, Kfir Aberman, Yael Pritch, and Daniel Cohen-Or.
\newblock Prompt-to-prompt image editing with cross attention control.
\newblock 2022.

\bibitem[Hertz et~al.(2023)Hertz, Aberman, and Cohen-Or]{hertz2023dds}
Amir Hertz, Kfir Aberman, and Daniel Cohen-Or.
\newblock Delta denoising score.
\newblock \emph{arXiv preprint arXiv:2304.07090}, 2023.

\bibitem[Ho et~al.(2020)Ho, Jain, and Abbeel]{ho2020ddpm}
Jonathan Ho, Ajay Jain, and Pieter Abbeel.
\newblock Denoising diffusion probabilistic models.
\newblock In \emph{Advances in Neural Information Processing Systems}, 2020.

\bibitem[Hsieh et~al.(2019)Hsieh, Chen, Chou, Shuai, Liu, and Cheng]{hsieh2019fashionon}
Chia-Wei Hsieh, Chieh-Yun Chen, Chien-Lung Chou, Hong-Han Shuai, Jiaying Liu, and Wen-Huang Cheng.
\newblock Fashionon: Semantic-guided image-based virtual try-on with detailed human and clothing information.
\newblock In \emph{Proceedings of the 27th ACM international conference on multimedia}, pp.\  275--283, 2019.

\bibitem[Hu et~al.(2021)Hu, Shen, Wallis, Allen-Zhu, Li, Wang, Wang, and Chen]{hu2021lora}
Edward~J Hu, Yelong Shen, Phillip Wallis, Zeyuan Allen-Zhu, Yuanzhi Li, Shean Wang, Lu~Wang, and Weizhu Chen.
\newblock Lora: Low-rank adaptation of large language models.
\newblock \emph{arXiv preprint arXiv:2106.09685}, 2021.

\bibitem[Huang et~al.(2024)Huang, He, Yu, Zhang, Si, Jiang, Zhang, Wu, Jin, Chanpaisit, et~al.]{huang2024vbench}
Ziqi Huang, Yinan He, Jiashuo Yu, Fan Zhang, Chenyang Si, Yuming Jiang, Yuanhan Zhang, Tianxing Wu, Qingyang Jin, Nattapol Chanpaisit, et~al.
\newblock Vbench: Comprehensive benchmark suite for video generative models.
\newblock In \emph{Proceedings of the IEEE/CVF Conference on Computer Vision and Pattern Recognition}, pp.\  21807--21818, 2024.

\bibitem[Issenhuth et~al.(2020)Issenhuth, Mary, and Calauzenes]{issenhuth2020not}
Thibaut Issenhuth, J{\'e}r{\'e}mie Mary, and Cl{\'e}ment Calauzenes.
\newblock Do not mask what you do not need to mask: a parser-free virtual try-on.
\newblock In \emph{Computer Vision--ECCV 2020: 16th European Conference, Glasgow, UK, August 23--28, 2020, Proceedings, Part XX 16}, pp.\  619--635. Springer, 2020.

\bibitem[Jiang et~al.(2020)Jiang, Zhang, Hong, Luo, Liu, and Bao]{jiang2020bcnet}
Boyi Jiang, Juyong Zhang, Yang Hong, Jinhao Luo, Ligang Liu, and Hujun Bao.
\newblock Bcnet: Learning body and cloth shape from a single image.
\newblock In \emph{European Conference on Computer Vision}, 2020.

\bibitem[Kawar et~al.(2022)Kawar, Zada, Lang, Tov, Chang, Dekel, Mosseri, and Irani]{kawar2022imagic}
Bahjat Kawar, Shiran Zada, Oran Lang, Omer Tov, Huiwen Chang, Tali Dekel, Inbar Mosseri, and Michal Irani.
\newblock Imagic: Text-based real image editing with diffusion models.
\newblock \emph{arXiv preprint arXiv:2210.09276}, 2022.

\bibitem[Kerbl et~al.(2023)Kerbl, Kopanas, Leimk{\"u}hler, and Drettakis]{kerbl20233d}
Bernhard Kerbl, Georgios Kopanas, Thomas Leimk{\"u}hler, and George Drettakis.
\newblock 3d gaussian splatting for real-time radiance field rendering.
\newblock \emph{ACM Transactions on Graphics (ToG)}, 42\penalty0 (4):\penalty0 1--14, 2023.

\bibitem[Kim et~al.(2024)Kim, Gu, Park, Park, and Choo]{kim2024stableviton}
Jeongho Kim, Guojung Gu, Minho Park, Sunghyun Park, and Jaegul Choo.
\newblock Stableviton: Learning semantic correspondence with latent diffusion model for virtual try-on.
\newblock In \emph{Proceedings of the IEEE/CVF Conference on Computer Vision and Pattern Recognition}, pp.\  8176--8185, 2024.

\bibitem[Kim \& Forsythe(2008)Kim and Forsythe]{kim2008adoption}
Jiyeon Kim and Sandra Forsythe.
\newblock Adoption of virtual try-on technology for online apparel shopping.
\newblock \emph{Journal of interactive marketing}, 22\penalty0 (2):\penalty0 45--59, 2008.

\bibitem[Kingma \& Welling(2013)Kingma and Welling]{kingma2013auto}
Diederik~P Kingma and Max Welling.
\newblock Auto-encoding variational bayes.
\newblock \emph{arXiv preprint arXiv:1312.6114}, 2013.

\bibitem[Kirillov et~al.(2023)Kirillov, Mintun, Ravi, Mao, Rolland, Gustafson, Xiao, Whitehead, Berg, Lo, et~al.]{kirillov2023segment}
Alexander Kirillov, Eric Mintun, Nikhila Ravi, Hanzi Mao, Chloe Rolland, Laura Gustafson, Tete Xiao, Spencer Whitehead, Alexander~C Berg, Wan-Yen Lo, et~al.
\newblock Segment anything.
\newblock In \emph{Proceedings of the IEEE/CVF International Conference on Computer Vision}, pp.\  4015--4026, 2023.

\bibitem[Lassner \& Zollhofer(2021)Lassner and Zollhofer]{lassner2021pulsar}
Christoph Lassner and Michael Zollhofer.
\newblock Pulsar: Efficient sphere-based neural rendering.
\newblock In \emph{Proceedings of the IEEE/CVF Conference on Computer Vision and Pattern Recognition}, pp.\  1440--1449, 2021.

\bibitem[Lee et~al.(2022)Lee, Gu, Park, Choi, and Choo]{lee2022high}
Sangyun Lee, Gyojung Gu, Sunghyun Park, Seunghwan Choi, and Jaegul Choo.
\newblock High-resolution virtual try-on with misalignment and occlusion-handled conditions.
\newblock In \emph{European Conference on Computer Vision}, pp.\  204--219. Springer, 2022.

\bibitem[Lerner et~al.(2007)Lerner, Govertsen, and McNellis]{lerner2007gaming}
Bill Lerner, Grant Govertsen, and Beth McNellis.
\newblock Gaming industry.
\newblock \emph{EPS (USD)}, 1\penalty0 (1.77):\penalty0 1--88, 2007.

\bibitem[Li et~al.(2023)Li, Li, Savarese, and Hoi]{li2023blip}
Junnan Li, Dongxu Li, Silvio Savarese, and Steven Hoi.
\newblock Blip-2: Bootstrapping language-image pre-training with frozen image encoders and large language models.
\newblock In \emph{International conference on machine learning}, pp.\  19730--19742. PMLR, 2023.

\bibitem[Li et~al.(2020)Li, Xu, Wei, and Yang]{li2020self}
Peike Li, Yunqiu Xu, Yunchao Wei, and Yi~Yang.
\newblock Self-correction for human parsing.
\newblock \emph{IEEE Transactions on Pattern Analysis and Machine Intelligence}, 2020.
\newblock \doi{10.1109/TPAMI.2020.3048039}.

\bibitem[Liu et~al.(2024)Liu, Zhang, Li, Yan, Gao, Chen, Yuan, Huang, Sun, Gao, et~al.]{liu2024sora}
Yixin Liu, Kai Zhang, Yuan Li, Zhiling Yan, Chujie Gao, Ruoxi Chen, Zhengqing Yuan, Yue Huang, Hanchi Sun, Jianfeng Gao, et~al.
\newblock Sora: A review on background, technology, limitations, and opportunities of large vision models.
\newblock \emph{arXiv preprint arXiv:2402.17177}, 2024.

\bibitem[Loper et~al.(2015)Loper, Mahmood, Romero, Pons-Moll, and Black]{SMPL:2015}
Matthew Loper, Naureen Mahmood, Javier Romero, Gerard Pons-Moll, and Michael~J. Black.
\newblock {SMPL}: A skinned multi-person linear model.
\newblock \emph{ACM Trans. Graphics, Asia}, 2015.

\bibitem[Ma et~al.(2024)Ma, Wang, Jia, Chen, Liu, Li, Chen, and Qiao]{ma2024latte}
Xin Ma, Yaohui Wang, Gengyun Jia, Xinyuan Chen, Ziwei Liu, Yuan-Fang Li, Cunjian Chen, and Yu~Qiao.
\newblock Latte: Latent diffusion transformer for video generation.
\newblock \emph{arXiv preprint arXiv:2401.03048}, 2024.

\bibitem[Men et~al.(2020)Men, Mao, Jiang, Ma, and Lian]{men2020controllable}
Yifang Men, Yiming Mao, Yuning Jiang, Wei-Ying Ma, and Zhouhui Lian.
\newblock Controllable person image synthesis with attribute-decomposed gan.
\newblock In \emph{Proceedings of the IEEE/CVF conference on computer vision and pattern recognition}, pp.\  5084--5093, 2020.

\bibitem[Meng et~al.(2010)Meng, Mok, and Jin]{meng2010interactive}
Yuwei Meng, Pik~Yin Mok, and Xiaogang Jin.
\newblock Interactive virtual try-on clothing design systems.
\newblock \emph{Computer-Aided Design}, 42\penalty0 (4):\penalty0 310--321, 2010.

\bibitem[Mildenhall et~al.(2020)Mildenhall, Srinivasan, Tancik, Barron, Ramamoorthi, and Ng]{mildenhall2020nerf}
Ben Mildenhall, Pratul~P Srinivasan, Matthew Tancik, Jonathan~T Barron, Ravi Ramamoorthi, and Ren Ng.
\newblock Nerf: Representing scenes as neural radiance fields for view synthesis.
\newblock In \emph{European Conference on Computer Vision}, 2020.

\bibitem[Mildenhall et~al.(2021)Mildenhall, Srinivasan, Tancik, Barron, Ramamoorthi, and Ng]{mildenhall2021nerf}
Ben Mildenhall, Pratul~P Srinivasan, Matthew Tancik, Jonathan~T Barron, Ravi Ramamoorthi, and Ren Ng.
\newblock Nerf: Representing scenes as neural radiance fields for view synthesis.
\newblock \emph{Communications of the ACM}, 2021.

\bibitem[Morelli et~al.(2023)Morelli, Baldrati, Cartella, Cornia, Bertini, and Cucchiara]{morelli2023ladi}
Davide Morelli, Alberto Baldrati, Giuseppe Cartella, Marcella Cornia, Marco Bertini, and Rita Cucchiara.
\newblock Ladi-vton: Latent diffusion textual-inversion enhanced virtual try-on.
\newblock \emph{arXiv preprint arXiv:2305.13501}, 2023.

\bibitem[Mou et~al.(2023)Mou, Wang, Xie, Zhang, Qi, Shan, and Qie]{mou2023t2i-adapter}
Chong Mou, Xintao Wang, Liangbin Xie, Jian Zhang, Zhongang Qi, Ying Shan, and Xiaohu Qie.
\newblock T2i-adapter: Learning adapters to dig out more controllable ability for text-to-image diffusion models.
\newblock \emph{arXiv preprint arXiv:2302.08453}, 2023.

\bibitem[Mystakidis(2022)]{mystakidis2022metaverse}
Stylianos Mystakidis.
\newblock Metaverse.
\newblock \emph{Encyclopedia}, 2022.

\bibitem[Pavlakos et~al.(2019)Pavlakos, Choutas, Ghorbani, Bolkart, Osman, Tzionas, and Black]{SMPL-X:2019}
Georgios Pavlakos, Vasileios Choutas, Nima Ghorbani, Timo Bolkart, Ahmed A.~A. Osman, Dimitrios Tzionas, and Michael~J. Black.
\newblock Expressive body capture: 3d hands, face, and body from a single image.
\newblock In \emph{IEEE Conference on Computer Vision and Pattern Recognition}, 2019.

\bibitem[Po et~al.(2024)Po, Yifan, Golyanik, Aberman, Barron, Bermano, Chan, Dekel, Holynski, Kanazawa, et~al.]{po2024state}
Ryan Po, Wang Yifan, Vladislav Golyanik, Kfir Aberman, Jonathan~T Barron, Amit Bermano, Eric Chan, Tali Dekel, Aleksander Holynski, Angjoo Kanazawa, et~al.
\newblock State of the art on diffusion models for visual computing.
\newblock In \emph{Computer Graphics Forum}, volume~43, pp.\  e15063. Wiley Online Library, 2024.

\bibitem[Pons-Moll et~al.(2017)Pons-Moll, Pujades, Hu, and Black]{pons2017clothcap}
Gerard Pons-Moll, Sergi Pujades, Sonny Hu, and Michael~J Black.
\newblock Clothcap: Seamless 4d clothing capture and retargeting.
\newblock \emph{ACM Transactions on Graphics (ToG)}, 36\penalty0 (4):\penalty0 1--15, 2017.

\bibitem[Radford et~al.(2021{\natexlab{a}})Radford, Kim, Hallacy, Ramesh, Goh, Agarwal, Sastry, Askell, Mishkin, Clark, et~al.]{radford2021clip}
Alec Radford, Jong~Wook Kim, Chris Hallacy, Aditya Ramesh, Gabriel Goh, Sandhini Agarwal, Girish Sastry, Amanda Askell, Pamela Mishkin, Jack Clark, et~al.
\newblock Learning transferable visual models from natural language supervision.
\newblock In \emph{ICML}, 2021{\natexlab{a}}.

\bibitem[Radford et~al.(2021{\natexlab{b}})Radford, Kim, Hallacy, Ramesh, Goh, Agarwal, Sastry, Askell, Mishkin, Clark, et~al.]{radford2021learning}
Alec Radford, Jong~Wook Kim, Chris Hallacy, Aditya Ramesh, Gabriel Goh, Sandhini Agarwal, Girish Sastry, Amanda Askell, Pamela Mishkin, Jack Clark, et~al.
\newblock Learning transferable visual models from natural language supervision.
\newblock In \emph{International Conference on Machine Learning}, 2021{\natexlab{b}}.

\bibitem[Ramamoorthi \& Hanrahan(2001)Ramamoorthi and Hanrahan]{ramamoorthi2001efficient}
Ravi Ramamoorthi and Pat Hanrahan.
\newblock An efficient representation for irradiance environment maps.
\newblock In \emph{Proceedings of the 28th annual conference on Computer graphics and interactive techniques}, pp.\  497--500, 2001.

\bibitem[Ramesh et~al.(2022)Ramesh, Dhariwal, Nichol, Chu, and Chen]{ramesh2022dalle2}
Aditya Ramesh, Prafulla Dhariwal, Alex Nichol, Casey Chu, and Mark Chen.
\newblock Hierarchical text-conditional image generation with clip latents.
\newblock \emph{arXiv preprint arXiv:2204.06125}, 2022.

\bibitem[Rombach et~al.(2022{\natexlab{a}})Rombach, Blattmann, Lorenz, Esser, and Ommer]{rombach2022high}
Robin Rombach, Andreas Blattmann, Dominik Lorenz, Patrick Esser, and Bj{\"o}rn Ommer.
\newblock High-resolution image synthesis with latent diffusion models.
\newblock In \emph{IEEE Conference on Computer Vision and Pattern Recognition}, 2022{\natexlab{a}}.

\bibitem[Rombach et~al.(2022{\natexlab{b}})Rombach, Blattmann, Lorenz, Esser, and Ommer]{rombach2022ldm}
Robin Rombach, Andreas Blattmann, Dominik Lorenz, Patrick Esser, and Bj{\"o}rn Ommer.
\newblock High-resolution image synthesis with latent diffusion models.
\newblock In \emph{IEEE Conference on Computer Vision and Pattern Recognition}, 2022{\natexlab{b}}.

\bibitem[Rong et~al.(2024)Rong, Grigorev, Wang, Black, Thomaszewski, Tsalicoglou, and Hilliges]{rong2024gaussian}
Boxiang Rong, Artur Grigorev, Wenbo Wang, Michael~J Black, Bernhard Thomaszewski, Christina Tsalicoglou, and Otmar Hilliges.
\newblock Gaussian garments: Reconstructing simulation-ready clothing with photorealistic appearance from multi-view video.
\newblock \emph{arXiv preprint arXiv:2409.08189}, 2024.

\bibitem[Ronneberger et~al.(2015)Ronneberger, Fischer, and Brox]{ronneberger2015u}
Olaf Ronneberger, Philipp Fischer, and Thomas Brox.
\newblock U-net: Convolutional networks for biomedical image segmentation.
\newblock In \emph{Medical Image Computing and Computer-Assisted Intervention--MICCAI 2015: 18th International Conference, Munich, Germany, October 5-9, 2015, Proceedings, Part III 18}, pp.\  234--241. Springer, 2015.

\bibitem[Ruiz et~al.(2023{\natexlab{a}})Ruiz, Li, Jampani, Pritch, Rubinstein, and Aberman]{ruiz2022dreambooth}
Nataniel Ruiz, Yuanzhen Li, Varun Jampani, Yael Pritch, Michael Rubinstein, and Kfir Aberman.
\newblock Dreambooth: Fine tuning text-to-image diffusion models for subject-driven generation.
\newblock In \emph{IEEE Conference on Computer Vision and Pattern Recognition}, 2023{\natexlab{a}}.

\bibitem[Ruiz et~al.(2023{\natexlab{b}})Ruiz, Li, Jampani, Pritch, Rubinstein, and Aberman]{ruiz2023dreambooth}
Nataniel Ruiz, Yuanzhen Li, Varun Jampani, Yael Pritch, Michael Rubinstein, and Kfir Aberman.
\newblock Dreambooth: Fine tuning text-to-image diffusion models for subject-driven generation.
\newblock In \emph{Proceedings of the IEEE/CVF Conference on Computer Vision and Pattern Recognition}, 2023{\natexlab{b}}.

\bibitem[Saharia et~al.(2022)Saharia, Chan, Saxena, Li, Whang, Denton, Ghasemipour, Gontijo~Lopes, Karagol~Ayan, Salimans, et~al.]{saharia2022photorealistic}
Chitwan Saharia, William Chan, Saurabh Saxena, Lala Li, Jay Whang, Emily~L Denton, Kamyar Ghasemipour, Raphael Gontijo~Lopes, Burcu Karagol~Ayan, Tim Salimans, et~al.
\newblock Photorealistic text-to-image diffusion models with deep language understanding.
\newblock \emph{Advances in neural information processing systems}, 35:\penalty0 36479--36494, 2022.

\bibitem[Shao et~al.(2023)Shao, Sun, Peng, Zheng, Zhou, Zhang, and Liu]{shao2023control4d}
Ruizhi Shao, Jingxiang Sun, Cheng Peng, Zerong Zheng, Boyao Zhou, Hongwen Zhang, and Yebin Liu.
\newblock Control4d: Dynamic portrait editing by learning 4d gan from 2d diffusion-based editor.
\newblock \emph{arXiv preprint arXiv:2305.20082}, 2023.

\bibitem[Singer et~al.(2023)Singer, Polyak, Hayes, Yin, An, Zhang, Hu, Yang, Ashual, Gafni, et~al.]{singer2022make-a-video}
Uriel Singer, Adam Polyak, Thomas Hayes, Xi~Yin, Jie An, Songyang Zhang, Qiyuan Hu, Harry Yang, Oron Ashual, Oran Gafni, et~al.
\newblock Make-a-video: Text-to-video generation without text-video data.
\newblock In \emph{International Conference on Learning Representations}, 2023.

\bibitem[Song et~al.(2021)Song, Meng, and Ermon]{song2021ddim}
Jiaming Song, Chenlin Meng, and Stefano Ermon.
\newblock Denoising diffusion implicit models.
\newblock In \emph{International Conference on Learning Representations}, 2021.

\bibitem[Stability.AI(2022)]{stable-diffusion}
Stability.AI.
\newblock Stable diffusion.
\newblock \url{https://stability.ai/blog/stable-diffusion-public-release}, 2022.

\bibitem[Stability.AI(2023{\natexlab{a}})]{deepfloyd-if}
Stability.AI.
\newblock Stability {AI} releases {D}eep{F}loyd {IF}, a powerful text-to-image model that can smartly integrate text into images.
\newblock \url{https://stability.ai/blog/deepfloyd-if-text-to-image-model}, 2023{\natexlab{a}}.

\bibitem[Stability.AI(2023{\natexlab{b}})]{stable-inpaint-diffusion}
Stability.AI.
\newblock Stable diffusion.
\newblock \url{https://huggingface.co/stabilityai/stable-diffusion-2-inpainting}, 2023{\natexlab{b}}.

\bibitem[Tang et~al.(2024)Tang, Ruiz, Chu, Li, Holynski, Jacobs, Hariharan, Pritch, Wadhwa, Aberman, et~al.]{tang2024realfill}
Luming Tang, Nataniel Ruiz, Qinghao Chu, Yuanzhen Li, Aleksander Holynski, David~E Jacobs, Bharath Hariharan, Yael Pritch, Neal Wadhwa, Kfir Aberman, et~al.
\newblock Realfill: Reference-driven generation for authentic image completion.
\newblock \emph{ACM Transactions on Graphics (TOG)}, 43\penalty0 (4):\penalty0 1--12, 2024.

\bibitem[Valevski et~al.(2022)Valevski, Kalman, Matias, and Leviathan]{valevski2022unitune}
Dani Valevski, Matan Kalman, Yossi Matias, and Yaniv Leviathan.
\newblock Unitune: Text-driven image editing by fine tuning an image generation model on a single image.
\newblock \emph{arXiv preprint arXiv:2210.09477}, 2022.

\bibitem[Vaswani et~al.(2017)Vaswani, Shazeer, Parmar, Uszkoreit, Jones, Gomez, Kaiser, and Polosukhin]{vaswani2017attention}
Ashish Vaswani, Noam Shazeer, Niki Parmar, Jakob Uszkoreit, Llion Jones, Aidan~N Gomez, {\L}ukasz Kaiser, and Illia Polosukhin.
\newblock Attention is all you need.
\newblock \emph{Advances in neural information processing systems}, 30, 2017.

\bibitem[Wang et~al.(2018)Wang, Zheng, Liang, Chen, Lin, and Yang]{wang2018toward}
Bochao Wang, Huabin Zheng, Xiaodan Liang, Yimin Chen, Liang Lin, and Meng Yang.
\newblock Toward characteristic-preserving image-based virtual try-on network.
\newblock In \emph{Proceedings of the European conference on computer vision (ECCV)}, pp.\  589--604, 2018.

\bibitem[Wang et~al.(2024)Wang, Cao, Han, and Wong]{wang2024survey}
Ruihe Wang, Yukang Cao, Kai Han, and Kwan-Yee~K Wong.
\newblock A survey on 3d human avatar modeling--from reconstruction to generation.
\newblock \emph{arXiv preprint arXiv:2406.04253}, 2024.

\bibitem[Wu et~al.(2022)Wu, Ge, Wang, Lei, Gu, Hsu, Shan, Qie, and Shou]{wu2022tuneavideo}
Jay~Zhangjie Wu, Yixiao Ge, Xintao Wang, Stan~Weixian Lei, Yuchao Gu, Wynne Hsu, Ying Shan, Xiaohu Qie, and Mike~Zheng Shou.
\newblock Tune-a-video: One-shot tuning of image diffusion models for text-to-video generation.
\newblock \emph{arXiv preprint arXiv:2212.11565}, 2022.

\bibitem[Wu et~al.(2024)Wu, Bian, Li, Wang, Reid, Torr, and Prisacariu]{wu2024gaussctrl}
Jing Wu, Jia-Wang Bian, Xinghui Li, Guangrun Wang, Ian Reid, Philip Torr, and Victor~Adrian Prisacariu.
\newblock Gaussctrl: Multi-view consistent text-driven 3d gaussian splatting editing.
\newblock \emph{arXiv preprint arXiv:2403.08733}, 2024.

\bibitem[Xie et~al.(2023)Xie, Huang, Dong, Zhao, Dong, Zhang, Zhu, and Liang]{xie2023gp}
Zhenyu Xie, Zaiyu Huang, Xin Dong, Fuwei Zhao, Haoye Dong, Xijin Zhang, Feida Zhu, and Xiaodan Liang.
\newblock Gp-vton: Towards general purpose virtual try-on via collaborative local-flow global-parsing learning.
\newblock In \emph{Proceedings of the IEEE/CVF Conference on Computer Vision and Pattern Recognition}, pp.\  23550--23559, 2023.

\bibitem[Xu et~al.(2024)Xu, Gu, Chen, and Chen]{xu2024ootdiffusion}
Yuhao Xu, Tao Gu, Weifeng Chen, and Chengcai Chen.
\newblock Ootdiffusion: Outfitting fusion based latent diffusion for controllable virtual try-on.
\newblock \emph{arXiv preprint arXiv:2403.01779}, 2024.

\bibitem[Yang et~al.(2023)Yang, Gu, Zhang, Zhang, Chen, Sun, Chen, and Wen]{yang2023paint}
Binxin Yang, Shuyang Gu, Bo~Zhang, Ting Zhang, Xuejin Chen, Xiaoyan Sun, Dong Chen, and Fang Wen.
\newblock Paint by example: Exemplar-based image editing with diffusion models.
\newblock In \emph{Proceedings of the IEEE/CVF Conference on Computer Vision and Pattern Recognition}, pp.\  18381--18391, 2023.

\bibitem[Zeng et~al.(2024)Zeng, Song, Nie, Tian, Wang, and Liu]{zeng2024cat}
Jianhao Zeng, Dan Song, Weizhi Nie, Hongshuo Tian, Tongtong Wang, and An-An Liu.
\newblock Cat-dm: Controllable accelerated virtual try-on with diffusion model.
\newblock In \emph{Proceedings of the IEEE/CVF Conference on Computer Vision and Pattern Recognition}, pp.\  8372--8382, 2024.

\bibitem[Zhang \& Agrawala(2023)Zhang and Agrawala]{zhang2023controlnet}
Lvmin Zhang and Maneesh Agrawala.
\newblock Adding conditional control to text-to-image diffusion models.
\newblock \emph{arXiv preprint arXiv:2302.05543}, 2023.

\bibitem[Zhang et~al.(2019)Zhang, Wang, Cao, and Wang]{zhang2019role}
Tingting Zhang, William Yu~Chung Wang, Ling Cao, and Yan Wang.
\newblock The role of virtual try-on technology in online purchase decision from consumers’ aspect.
\newblock \emph{Internet Research}, 29\penalty0 (3):\penalty0 529--551, 2019.

\bibitem[Zhao et~al.(2023)Zhao, Li, Hu, Li, Zou, Shi, and Fan]{zhao2023t2p}
Rui Zhao, Wei Li, Zhipeng Hu, Lincheng Li, Zhengxia Zou, Zhenwei Shi, and Changjie Fan.
\newblock Zero-shot text-to-parameter translation for game character auto-creation.
\newblock In \emph{IEEE Conference on Computer Vision and Pattern Recognition}, 2023.

\bibitem[Zhou et~al.(2024{\natexlab{a}})Zhou, Zhou, Cheng, Feng, and Hou]{zhou2024storydiffusion}
Yupeng Zhou, Daquan Zhou, Ming-Ming Cheng, Jiashi Feng, and Qibin Hou.
\newblock Storydiffusion: Consistent self-attention for long-range image and video generation.
\newblock \emph{arXiv preprint arXiv:2405.01434}, 2024{\natexlab{a}}.

\bibitem[Zhou et~al.(2024{\natexlab{b}})Zhou, Ma, Fan, and Yang]{zhou2024headstudio}
Zhenglin Zhou, Fan Ma, Hehe Fan, and Yi~Yang.
\newblock Headstudio: Text to animatable head avatars with 3d gaussian splatting.
\newblock \emph{arXiv preprint arXiv:2402.06149}, 2024{\natexlab{b}}.

\bibitem[Zhu et~al.(2023)Zhu, Yang, Zhu, Reda, Chan, Saharia, Norouzi, and Kemelmacher-Shlizerman]{zhu2023tryondiffusion}
Luyang Zhu, Dawei Yang, Tyler Zhu, Fitsum Reda, William Chan, Chitwan Saharia, Mohammad Norouzi, and Ira Kemelmacher-Shlizerman.
\newblock Tryondiffusion: A tale of two unets.
\newblock In \emph{Proceedings of the IEEE/CVF Conference on Computer Vision and Pattern Recognition}, pp.\  4606--4615, 2023.

\bibitem[Zhuang et~al.(2023)Zhuang, Wang, Liu, Lin, and Li]{zhuang2023dreameditor}
Jingyu Zhuang, Chen Wang, Lingjie Liu, Liang Lin, and Guanbin Li.
\newblock Dreameditor: Text-driven 3d scene editing with neural fields.
\newblock \emph{arXiv preprint arXiv:2306.13455}, 2023.

\bibitem[Zhuang et~al.(2024)Zhuang, Kang, Cao, Li, Lin, and Shan]{zhuang2024tip}
Jingyu Zhuang, Di~Kang, Yan-Pei Cao, Guanbin Li, Liang Lin, and Ying Shan.
\newblock Tip-editor: An accurate 3d editor following both text-prompts and image-prompts.
\newblock \emph{arXiv preprint arXiv:2401.14828}, 2024.

\end{thebibliography}
\bibliographystyle{conference}

\clearpage
\appendix

\section{Analysis of the number of views used in reference-driven image editing}
\label{sec:number-views}
As we have discussed in Sec.~\ref{sec:personalized_inpaiting} and Sec.~\ref{sec:3dgs_editing}, we first achieve personalized diffusion model by training LoRA on an edited training image set $X_{\text{train}}$ that contains 4 images in our experiments. We subsequently employed this fine-tuned model to perform 3DGS editing, enhancing multi-view consistency through our proposed persona-aware 3DGS editing methodology.

\begin{figure}[h]
  \centering
   \includegraphics[width=1\linewidth]{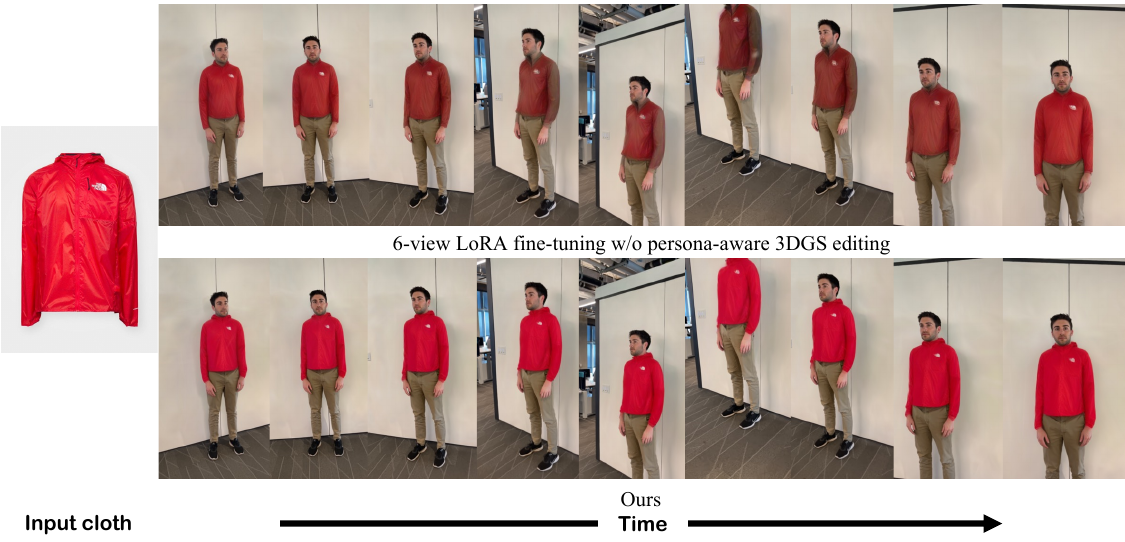}   
  \caption{\textbf{Effectiveness of persona-aware 3DGS editing while using more views for LoRA fine-tuning.} }
   \label{fig:analysis-views}
\end{figure}

It is important to address the necessity of our persona-aware approach, as one might assume that a personalized diffusion model could be further improved by incorporating additional training images. To elucidate the significance of persona-aware 3DGS editing, we conducted ablation studies using a six-image training set while disabling the persona-aware component. The results, illustrated in Fig.~\ref{fig:analysis-views}, demonstrate that even with a greater number of views, relying solely on the diffusion model for 3DGS scene editing leads to persistent inconsistency issues. In contrast, our persona-aware 3DGS editing effectively resolves these challenges, enhancing multi-view coherence. 

Moreover, increasing the number of training views imposes higher demands on GPU memory and adversely impacts training efficiency. Due to these GPU memory constraints, we are limited to experiments with no more than six views.

\section{Additional Qualitative Comparison}
\begin{figure}[t]
  \centering
   \includegraphics[width=\linewidth]{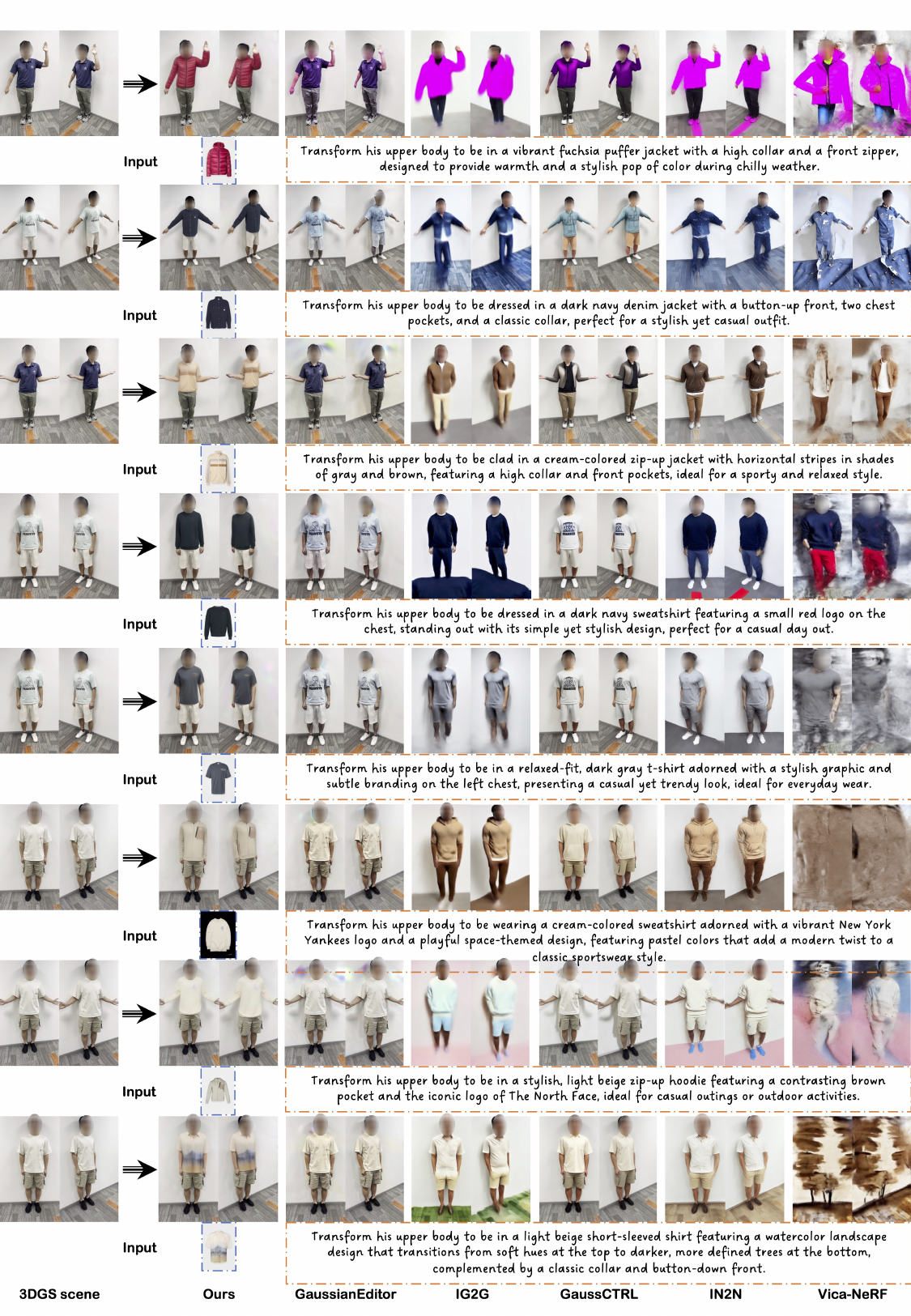}
  \caption{\textbf{Additional qualitative comparison with existing 3D scene editing techniques.}}
   \label{fig:more-comparison}
\end{figure}

We provide more qualitative comparisons with four baseline methods~\citep{chen2023gaussianeditor, ig2g, wu2024gaussctrl, haque2023instructnerf2nerf, dong2024vica} in Fig.~\ref{fig:more-comparison}.
These results reinforce the claims made in Sec.~\ref{sec:qualitative} of the main paper, providing further evidence of the superior performance of our \OMO in achieving high-fidelity 3D virtual try-on. 
The results demonstrate our method’s capability to produce realistic textures and overall high visual quality that accurately represent garment characteristics, establishing it as a state-of-the-art solution for this task.

To further enhance the immersive experience and understanding of our results, we include multiple outcomes of our \OMO and qualitative comparisons as \textbf{$360^{\circ}$ rotating videos}. These videos are accessible via the \textbf{HTML file} in the supplementary zip file, allowing viewers to explore the generated avatars from various angles and perspectives. Additionally, we provide a demo video in the supplementary zip file for improved visualization.

\clearpage
\section{Illustration of User Studies}

We conducted the user studies in the form of a questionnaire supported by Google Forms. Each volunteer will be presented with 25 randomly selected generated results with rotating videos, with the description to be:

``You will be presented with 25 sets of results with each generated by 6 3D scene editing methods. Your task is to evaluate and mark the best results in terms of (1) most realistic after try-on, (2) similarity with the cloth image , and (3) overall performance. Note that the text is supposed to describe the cloth image''
\begin{figure}[h]
  \centering
   \includegraphics[width=1\linewidth]{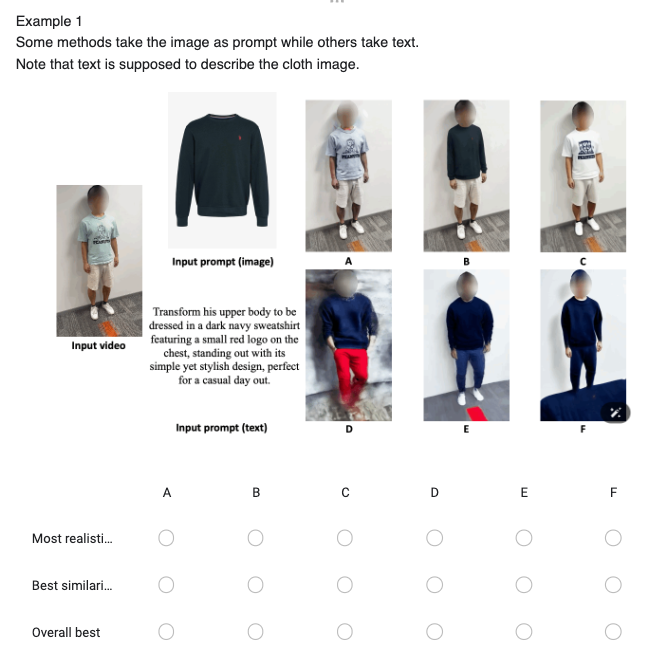}   
  \caption{\textbf{ An example of our user studies via Google Form. }  Note that all the results shown here are in the format of rotating videos.}
   \label{fig:user-studies-screenshot}
\end{figure} 

\clearpage
\section{Societal Impact}

The development of 3D virtual try-on technology has vast applications in AR/VR environments and online shopping experiences. However, it also raises significant concerns regarding potential malicious use. First, privacy issues are increasingly critical, as these systems often require the collection of sensitive personal data, including body measurements and shopping preferences. If this data is inadequately protected, it can be exploited by malicious actors, resulting in identity theft or targeted cyberattacks. Second, there is a risk of deepfake applications, where malicious users could manipulate the technology to create hyper-realistic, altered images or videos that misrepresent individuals, potentially leading to harassment, defamation, or misinformation. These risks highlight the urgent need for robust safeguards to prevent the malicious use of virtual try-on technology, thereby protecting individuals and preserving the integrity of digital interactions.

\end{document}